\documentclass[10pt,twocolumn,letterpaper]{article}

\usepackage[pagenumbers]{cvpr} 

\usepackage{graphicx}
\usepackage{amsmath}
\usepackage{amssymb}
\usepackage{booktabs}

\usepackage[pagebackref,breaklinks,colorlinks]{hyperref}

\usepackage[capitalize]{cleveref}
\crefname{section}{Sec.}{Secs.}
\Crefname{section}{Section}{Sections}
\Crefname{table}{Table}{Tables}
\crefname{table}{Tab.}{Tabs.}

\usepackage[accsupp]{axessibility} 
\usepackage{graphicx}
\usepackage{tikz}
\usepackage{subcaption}
\usepackage{multirow}

\def\rvx{{\mathbf{x}}}

\def\rvz{{\mathbf{z}}}
\def\N{{\mathcal{N}}}
\def\E{{\mathbb{E}}}
\def\KL{{\mathrm{KL}}}
\def\cvprPaperID{8231} 
\def\confName{CVPR}
\def\confYear{2023}
\usepackage[ruled,linesnumbered]{algorithm2e}
\newcommand\blfootnote[1]{%
  \begingroup
  \renewcommand\thefootnote{}\footnote{#1}%
  \addtocounter{footnote}{-1}%
  \endgroup
}
\begin{document}

\title{Learning Joint Latent Space EBM Prior Model for Multi-layer Generator}

\author{Jiali Cui$^1$, Ying Nian Wu$^2$, Tian Han$^1$\thanks{: corresponding author}\\
$^1$Department of Computer Science, Stevens Institute of Technology\\
$^2$Department of Statistics, University of California, Los Angeles\\
{\tt\small \{jcui7,than6\}@stevens.edu, ywu@stat.ucla.edu}
}

\maketitle
\renewcommand*{\thefootnote}{\arabic{footnote}}
\begin{abstract}
This paper studies the fundamental problem of learning multi-layer generator models. The multi-layer generator model builds multiple layers of latent variables as a prior model on top of the generator, which benefits learning complex data distribution and hierarchical representations. However, such a prior model usually focuses on modeling inter-layer relations between latent variables by assuming non-informative (conditional) Gaussian distributions, which can be limited in model expressivity. To tackle this issue and learn more expressive prior models, we propose an energy-based model (EBM) on the joint latent space over all layers of latent variables with the multi-layer generator as its backbone. Such joint latent space EBM prior model captures the intra-layer contextual relations at each layer through layer-wise energy terms, and latent variables across different layers are jointly corrected. We develop a joint training scheme via maximum likelihood estimation (MLE), which involves Markov Chain Monte Carlo (MCMC) sampling for both prior and posterior distributions of the latent variables from different layers. To ensure efficient inference and learning, we further propose a variational training scheme where an inference model is used to amortize the costly posterior MCMC sampling. Our experiments demonstrate that the learned model can be expressive in generating high-quality images and capturing hierarchical features for better outlier detection. 

\end{abstract}

\section{Introduction}
Deep generative models (a.k.a, \textit{generator models}) have made promising progress in learning complex data distributions and achieved great successes in image and video synthesis \cite{karras2019style,saito2020train, song2020score,Tulyakov_2018_CVPR} as well as representation learning \cite{zhao2017learning,DBLP:conf/iclr/Child21}. Such models usually consist of low-dimensional latent variables together with a top-down generation model that maps such latent factors to the observed data. The latent factors can serve as an abstract data representation, but it is often modelled via a single latent vector with non-informative prior distribution which leads to limited model expressivity and fails to capture different levels of abstractions. Learning an informative prior model for hierarchical representations is needed, yet research in this direction is still under-developed.

A principled way to learn such a prior model is by learning the generator models with multiple layers of latent variables. However, the learning of multi-layer generator model can be challenging as the inter-layer structural relation (i.e., latent variables across different layers) and the intra-layer contextual relation (i.e., latent units within the same layer) have to be effectively modelled and efficiently learned. Various methods have been proposed \cite{NIPS2016_6ae07dcb,nijkamp2020learning,vahdat2020nvae,maaloe2019biva,DBLP:conf/iclr/Child21}, but they only focused on inter-layer modeling by assuming the conditional Gaussian distribution across different layers while ignoring the intra-layer contextual modeling as the latent units are \textit{conditional independent} within each layer. 

The energy-based models (EBMs), on the other hand, are shown to be expressive and proved to be powerful in capturing contextual and non-structural data regularities. Notably, \cite{pang2020learning} considers the EBM in the latent space for the non-hierarchical generator model, where the energy function is considered as a correction of the non-informative Gaussian prior. The low dimensionality of the latent space makes EBM effective in capturing regularities in the data. However, a single latent vector in \cite{pang2020learning} is infeasible for capturing the patterns at multiple layers of abstractions, which limits its model capacity.

In this paper, we propose to combine the strengths of the latent space EBM and the generator with multiple layers of latent variables for better hierarchical representations and a more expressive prior model. Specifically, we introduce layer-wise energy terms to exponentially tilt the non-informative Gaussian conditional at each layer, and latent variables across different layers are modelled jointly through EBM with the multi-layer generator model as its backbone. Such a joint EBM prior model seamlessly integrates the intra-layer contextual modeling via layer-wise energy terms and inter-layer structural modeling with multi-layer latent variables. 

The joint EBM prior model can be learned by maximum likelihood estimation (MLE). Each learning iteration involves Markov chain Monte Carlo (MCMC) sampling of latent variables in each layer from both the prior and posterior distributions. The prior sampling can be efficiently done due to the low dimensionality of the latent variables and, more importantly, the lightweight networks for energy functions, while the posterior sampling can be less efficient. Therefore, we further develop the variational training scheme where an additional inference model is used for posterior approximation and is jointly trained with the joint EBM prior model. 

\noindent \textbf{Contributions: }1) We propose a joint latent space EBM prior model for the generator model with multiple layers of latent variables; 2) We develop the maximum likelihood learning algorithm that learns the joint EBM prior model based on MCMC prior and posterior sampling across different layers. We further propose the variational joint training scheme for efficient learning and inference; 3) We provide strong empirical results through extensive experiments. 

\section{Background}
In this section, we present the background of multi-layer latent variable model and latent space EBM prior model, which shall serve as the foundation of the proposed model.
\subsection{Multi-layer latent variable model}
\label{section:HVAE}
Let $\rvx$ be the high-dimensional observed example, and $\rvz$ be the low-dimensional latent variables. The latent variable generative model, or \textit{generator model}, factorizes a joint distribution of $(\rvx, \rvz)$ as
\begin{eqnarray}
\label{latent-generative-model}
     p_{\beta}(\rvx,\rvz) = p_{\beta_0}(\rvx|\rvz)p_{\beta_{>0}}(\rvz) 
\end{eqnarray}
where $p_{\beta_0}(\rvx|\rvz)$ is the generation model with parameter $\beta_0$ that maps from latent space to data space, and $p_{\beta_{>0}}(\rvz)$ is the prior distribution over latent variables with parameter $\beta_{>0}$. $\beta=\{\beta_0, \beta_{>0}\}$.

\noindent \textbf{Gaussian prior model:} For non-hierarchical models \cite{kingma2013auto,goodfellow2014generative}, $p_{\beta_{>0}}(\rvz)$ is defined on single layer of latent variables and is typically assumed to be uniform or unit Gaussian. For hierarchical models with multiple layers of latent variables \cite{NIPS2016_6ae07dcb,nijkamp2020learning}, $p_{\beta_{>0}}(\rvz)$ can be further decomposed into conditional distributions between consecutive layers of latent variables as
\begin{eqnarray}
\label{gaussian-pz}
     p_{\beta_>{0}}(\rvz) = \prod_{i=1}^{L-1}p_{\beta_i}(\rvz_i|\rvz_{i+1})p(\rvz_{L})
\end{eqnarray}
where $p_{\beta_i}(\rvz_i|\rvz_{i+1}) \sim \N (\mu_{\beta_i}(\rvz_{i+1}), \sigma^2_{\beta_i}(\rvz_{i+1}))$ and is parameterized by a network with parameter $\beta_i$, and $p(\rvz_L)$ is chosen to be a simple distribution, such as uniform or unit Gaussian. 

\noindent \textbf{Maximum likelihood learning:}
Learning such latent variable generative models can be done using maximum likelihood estimation (MLE). The marginal distribution is $p_{\beta}(\rvx) = \int p_{\beta}(\rvx, \rvz)d\rvz$ with the gradient:
\begin{eqnarray}
\label{vae-mle}
    \nabla_{\beta}\log p_{\beta}(\rvx) = \E_{p_{\beta}(\rvz|\rvx)}[\nabla_{\beta}\log p_{\beta}(\rvx, \rvz)]
\end{eqnarray}
where the expectation can be approximated via Monte Carlo sampling from the posterior distribution $p_{\beta}(\rvz|\rvx)$. The MLE can then be accomplished through gradient ascent using such gradients. The posterior sampling usually requires the Markov Chain Monte Carlo (MCMC) such as Langevin dynamics \cite{han2017alternating,nijkamp2020learning}

\noindent \textbf{Variational learning:} To alleviate the computational burden of MCMC, variational approach \cite{DBLP:journals/corr/BurdaGS15} introduces an additional inference model $q_{\omega}(\rvz|\rvx)$ with a separate set of parameters $\omega=(\omega_1, \dots, \omega_L)$ for posterior approximation, 
\begin{eqnarray}
\label{gaussian-qz}
     q_{\omega}(\rvz|\rvx) = q_{\omega_1}(\rvz_{1}|\rvx)\prod_{i=1}^{L-1}q_{\omega_{i+1}}(\rvz_{i+1}|\rvz_{i})
\end{eqnarray}
where $q_{\omega_1}(\rvz_1|\rvx)$ and $q_{\omega_{i+1}}(\rvz_{i+1}|\rvz_i)$ are usually assumed as conditional Gaussian distributions, forming a ``bottom-up'' inference structure. The generator and inference model can be jointly learned via maximizing the evidence lower bound (ELBO), i.e., $\max_{\beta, \omega}\text{ELBO}(\beta, \omega)$, where ELBO is defined as $\text{ELBO}(\beta, \omega)
=\E_{q_{\omega}(\rvz|\rvx)} [\log p_{\beta_0}(\rvx|\rvz)] - D_{\KL}(q_{\omega}(\rvz|\rvx)||p_{\beta_{>0}}(\rvz))$. 

\subsection{Latent space energy-based model}
The energy-based model (EBM) offers a flexible approach for learning the data distribution and is shown to be expressive in capturing data regularities \cite{nijkamp2019learning,du2019implicit,du2020improved,yin2020analyzing,gao2020learning,xiao2020vaebm}. Most existing works focus on learning the EBM on data space, which is high-dimensional and can be challenging. To tackle this challenge, \cite{pang2020learning,aneja2021contrastive,cui2023learning} propose to learn latent space EBM as an informative prior model. With low-dimensional latent space, learning the EBM can be more efficient and effective, which in turn benefits the expressivity of the whole model. Specifically, \cite{pang2020learning} considers the latent space energy-based prior model on a \textit{single layer} of latent variables, 
\begin{eqnarray}
\label{lebm}
     p_{\alpha}(\rvz) = \frac{1}{\mathrm{Z}(\alpha)}\exp{[f_{\alpha}(\rvz)]}p_0(\rvz)
\end{eqnarray}
where $-f_{\alpha}(\rvz)$ is the energy function, $\mathrm{Z}(\alpha)$ is the normalizing constant, i.e., $\mathrm{Z}(\alpha)=\int \exp{[f_{\alpha}(\rvz)]}p_0(\rvz)d\rvz$, and $p_0(\rvz)$ is the reference distribution assumed to be unit Gaussian. Compared to data space EBMs in which the energy function needs to support the entire high-dimensional space, such exponential tilting latent space EBMs can be more efficient in capturing data regularities.

\section{Model and Learning}
\subsection{Joint latent space EBM prior model}
\label{section:Our-model}
\begin{figure}[t]
\centering
\includegraphics[width=.95\columnwidth]{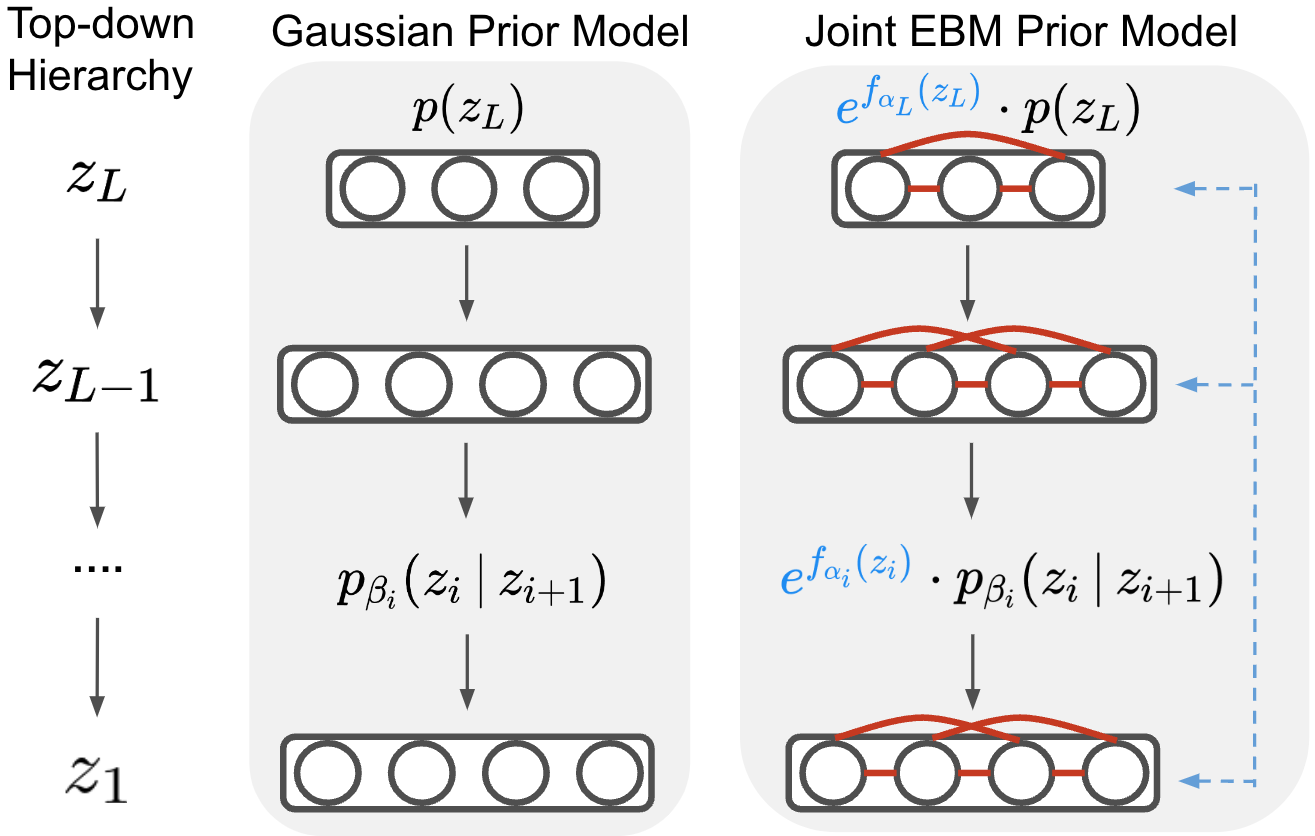}
\caption{\textbf{Left panel:} Gaussian prior model. \textbf{Right panel:} Joint EBM prior model. \textbf{Black solid lines with arrow:} inter-layer relations modelling. \textbf{Red solid lines}: intra-layer contextual relations modelling. \textbf{Blue dashed lines:} joint modelling upon all layers. }
\label{Fig.illustration}
\end{figure}

For generator models with multi-layer latent variables (or \textit{multi-layer generator model}), consecutive layers are modelled by conditional Gaussian distributions (see Eqn.\ref{gaussian-pz}), which essentially assumes the \textit{conditional independence} for latent units within the $i$-th layer given the $(i+1)$-th layer of latent variables. Such a conditional independence assumption limits the model capacity as the contextual relation between latent units within each layer is largely ignored (see Fig.\ref{Fig.illustration}), and needs to be improved for informative conditional modeling and better model expressivity. In this paper, we propose the joint EBM prior for multi-layer generator models, 
\begin{eqnarray}
\label{our-prior-model}
     &&p_{\alpha, \beta_{>0}}(\rvz) = \frac{1}{\mathrm{Z}_{\alpha, \beta_{>0}}}\exp{[f_{\alpha}(\rvz)]}p_{\beta_{>0}}(\rvz) \\
     &=& \frac{1}{\mathrm{Z}_{\alpha, \beta_{>0}}}\exp\left[\sum_{i=1}^{L}f_{\alpha_i}(\rvz_i)\right]\prod_{i=1}^{L-1}p_{\beta_i}(\rvz_i|\rvz_{i+1})p(\rvz_{L})  \nonumber
\end{eqnarray}
where we denote $\alpha=(\alpha_1, ..., \alpha_L)$ for EBM parameters and $\rvz = (\rvz_1, ..., \rvz_L)$ for latent variables in different layers with layer $L$ being the top layer. $\mathrm{Z}_{\alpha, \beta_{>0}} = \int \exp{[f_{\alpha}(\rvz)]}p_{\beta_{>0}}(\rvz) d\rvz$ is the normalizing constant regarding latent variables for all layers. Thus, the latent variables across different layers are \textit{jointly} corrected via EBM prior as in Eqn.\ref{our-prior-model}, where $f_{\alpha}(\rvz)$ is the energy function for latent variables from all layers. 

In this paper, we consider a simple factorized layer-wise parameterization, i.e., $f_{\alpha}(\rvz)= \sum_{i=1}^{L}f_{\alpha_i}(\rvz_i)$, but other parameterizations are also feasible, which we will explore in future work. With such energy parameterization, it's worth noting that the \textit{un-normalized} prior model can be viewed as layer-wise exponential tilting,
\begin{equation}\label{our-prior-model-2}
\begin{aligned} 
&\underbrace{\exp\left[\sum_{i=1}^{L}f_{\alpha_i}(\rvz_i)\right]}_{\text{Energy correction}}\underbrace{\prod_{i=1}^{L-1}p_{\beta_i}(\rvz_i|\rvz_{i+1})p(\rvz_{L})}_{\text{Gaussian prior}} \\
=&\underbrace{\exp\left[f_{\alpha_L}(\rvz_L)\right]p(\rvz_{L})}_{\text{Correction on top layer}}\prod_{i=1}^{L-1}\underbrace{\exp{[f_{\alpha_i}(\rvz_i)]}p_{\beta_i}(\rvz_i|\rvz_{i+1})}_{\text{Correction on intermediate layer}}
\end{aligned}
\end{equation}
See Fig.\ref{Fig.illustration} for an illustration and comparison with multi-layer generator model with Gaussian prior.

\noindent \textbf{Joint vs. conditional EBM prior:} Besides the proposed joint modeling, it is also tempting to consider EBM prior for layer-wise Gaussian conditional, i.e., $\tilde{p}_{\alpha_i, \beta_i}(\rvz_i|\rvz_{i+1})=\frac{1}{\mathrm{Z}(\rvz_{i+1})}\exp{[f_{\alpha_i}(\rvz_i)]}p_{\beta_i}(\rvz_i|\rvz_{i+1})$, and form the overall prior $p_{\alpha, \beta_{>0}}(\rvz) = \prod_{i=1}^L\tilde{p}_{\alpha_i,\beta_i}(\rvz_i|\rvz_{i+1})p(\rvz_{L})$. Such a scheme is closely related to autoregressive energy machine \cite{DBLP:conf/icml/DurkanN19} and is adopted in NCP-VAE \cite{aneja2021contrastive}. However, the normalizing constant $\mathrm{Z}(\rvz_{i+1})$ in $\tilde{p}_{\alpha_i, \beta_i}(\rvz_i|\rvz_{i+1})$ involves the latent variable $\rvz_{i+1}$ from the upper layer which can be \textit{intractable} and needs an additional inner-loop for sampling or optimization. The proposed joint EBM prior 
couples the latent variables across different layers via energy function and can be learned effectively and efficiently.

\subsection{Maximum Likelihood Estimation}
\label{section:MLE}
Our joint EBM prior model can be trained using MLE. Let $\theta = (\alpha, \beta)$ denotes the model parameters and $\theta$ can be learned by maximizing the log-likelihood on $n$ training observations
\begin{equation}
\label{marginal}
\begin{aligned} 
L(\theta) &= \sum_{i=1}^{n}\log p_{\theta}(\rvx_i)=\sum_{i=1}^{n}\log \int p_{\beta_0}(\rvx|\rvz)p_{\alpha, \beta_{>0}}(\rvz)d\rvz \nonumber
\end{aligned}
\end{equation}
When $n$ becomes sufficiently large, maximizing the above log-likelihood is equivalent to minimizing the Kullback-Leibler (KL) divergence between model distribution and empirical data distribution, i.e., $\min_\theta D_{\mathrm{KL}}(p_{\rm data}(\rvx)||p_{\theta}(\rvx))$. 

To update the parameter $\theta$, we can compute the the gradient of log-likelihood $\nabla_{\theta}\log p_{\theta}(\rvx)$ as
\begin{equation}
\label{mle-joint}
\begin{aligned} 
 \nabla_{\theta}\log p_{\theta}(\rvx) &= \E_{p_{\theta}(\rvz|\rvx)}[\nabla_{\theta}\log p_{\theta}(\rvx, \rvz)] \\
 &= \E_{p_{\theta}(\rvz|\rvx)}[\nabla_{\theta}\log p_{\beta_0}(\rvx|\rvz)] \\
 &+ \E_{p_{\theta}(\rvz|\rvx)}[\nabla_{\theta}\log p_{\alpha, \beta_{>0}}(\rvz)]
\end{aligned}
\end{equation}
With such a gradient, we can learn $\theta$ using gradient ascent. 

\noindent \textbf{Learning generation model $\beta_0$:} $p_{\beta_0}(\rvx|\rvz)$ is assumed to be Gaussian distribution, i.e.,  $p_{\beta_0}(\rvx|\rvz) \sim \N(g_{\beta_0}(\rvz), \sigma^2I)$, with generation network $g_{\beta_0}$ with parameter $\beta_0$ and pre-specified $\sigma^2$ for simplicity. The learning gradient $\nabla_{\beta_0}\log p_{\theta}(\rvx)$ can then be expressed as
\begin{eqnarray}
 \label{x-generator-grad}
     \nabla_{\beta_0}\log p_{\theta}(\rvx) &=& \E_{p_{\theta}(\rvz|\rvx)}[\nabla_{\beta_0}\log p_{\beta_0}(\rvx|\rvz)] \\
     &=& \E_{p_{\theta}(\rvz|\rvx)}\left[ - \nabla_{\beta_0}\frac{||\rvx - g_{\beta_0}(\rvz)||^2}{2\sigma^2}\right] \nonumber
\end{eqnarray}

\noindent \textbf{Learning prior model $\alpha$, $\beta_{>0}$:} Learning $\alpha_i$ can be done by computing the gradient $\nabla_{\alpha_i}\log p_{\theta}(\rvx)$ as
\begin{eqnarray}
 \label{energy-grad}
\nabla_{\alpha_i}\log p_{\theta}(\rvx)  &=& \E_{p_{\theta}(\rvz|\rvx)}[\nabla_{\alpha_i}f_{\alpha_i}(\rvz_i)] \\
&-&\E_{p_{\alpha, \beta_{>0}}(\rvz)}[\nabla_{\alpha_i}f_{\alpha_i}(\rvz_i)] \nonumber
\end{eqnarray}
For updating $\beta_{>0}$, the gradient $\nabla_{\beta_{i}}\log p_{\theta}(\rvx)$ is
\begin{eqnarray}
 \label{z-generator-grad}
\nabla_{\beta_i}\log p_{\theta}(\rvx)  &=&\E_{p_{\theta}(\rvz|\rvx)}[\nabla_{\beta_i}\log p_{\beta_i}(\rvz_i|\rvz_{i+1})] \\
&-& \E_{p_{\alpha, \beta_{>0}}(\rvz)}[\nabla_{\beta_i}\log p_{\beta_i}(\rvz_i|\rvz_{i+1})] \nonumber
\end{eqnarray}

\noindent \textbf{Sampling:} Both Eqn.\ref{energy-grad} and Eqn.\ref{z-generator-grad} require sampling from the posterior and prior distribution, which can be done via Langevin dynamic (LD) \cite{lemons1997paul}. Given a target distribution $p(\rvz)$, Langevin dynamic samples $\rvz \sim p(\rvz)$ by computing the gradient $\nabla_\rvz \log p(\rvz)$ and iteratively update $\rvz$ as
\begin{eqnarray}
\label{langevin}
\rvz_t = \rvz_{t-1} + s\nabla_\rvz \log p(\rvz_{t-1}) + \sqrt{2s}\epsilon_{t-1} 
\end{eqnarray}
where $t$ indexes the time step, $s$ is the step size, and $\epsilon$ is the Gaussian noise for each time step.

\noindent \textbf{Prior sampling:} By replacing target $p(\rvz)$ with $p_{\alpha, \beta_{>0}}(\rvz)$, the prior sampling computes $\nabla_\rvz \log p_{\alpha, \beta_{>0}}(\rvz)$ as
\begin{eqnarray}
\label{prior-langevin}
\nabla_\rvz \left[\sum_{i=1}^{L}f_{\alpha_i}(\rvz_i)+ \sum_{i=1}^{L-1}\log p_{\beta_i}(\rvz_i|\rvz_{i+1}) + \log p(\rvz_{L})\right]
\end{eqnarray}
\noindent \textbf{Posterior sampling:}  By replacing $p(\rvz)$ with $p_{\theta}(\rvz|\rvx)$, where $p_{\theta}(\rvz|\rvx) \propto p_{\beta_0}(\rvx|\rvz)p_{\alpha, \beta_{>0}}(\rvz)$, the posterior sampling computes $\nabla_\rvz \log p_{\theta}(\rvz|\rvx)$ as
\begin{eqnarray}
\label{posterior-langevin}
\nabla_{\rvz}\log p_{\theta}(\rvz|\rvx) =\nabla_\rvz [\log p_{\beta_0}(\rvx|\rvz) + \log p_{\alpha, \beta_{>0}}(\rvz)]
\end{eqnarray}
Notice that posterior sampling can be computationally inefficient as $\nabla_\rvz [\log p_{\beta_0}(\rvx|\rvz)$] requires back-propagation through the deep generation model. 



\subsection{Variational Learning}
For efficient posterior sampling, an inference model $q_\omega(\rvz|\rvx)$ with a separate set of parameters $\omega$ can be used for the posterior approximation. In this paper, we use the bottom-up inference model as Eqn.\ref{gaussian-qz} for amortizing the costly posterior MCMC sampling. Particularly, instead of KL minimization between marginal distributions as in MLE (see Sec.\ref{section:MLE}), we consider the KL optimization between two joint densities, one for generator model density, i.e., $p_\theta(\rvx, \rvz)=p_{\beta_0}(\rvx|\rvz)p_{\alpha,\beta_{>0}}(\rvz)$, and one for data density, i.e., $q_\omega(\rvx, \rvz) = p_{\rm data}(\rvx)q_\omega(\rvz|\rvx)$. We propose joint learning through KL minimization, denoting the objective to be $L(\theta, \omega)$, i.e., 
\begin{eqnarray}\label{eq:kl_joint}
\min_\theta \min_\omega L(\theta, \omega) = \min_\theta \min_\omega D_{\mathrm{KL}}(q_{\omega}(\rvx,\rvz)||p_{\theta}(\rvx,\rvz))
\end{eqnarray}

\noindent \textbf{Learning generation model $\beta_0$:} For learning $\beta_0$, we can compute the gradient as
\begin{eqnarray}
 \label{x-generator-grad-2}
     -\nabla_{\beta_0}L(\theta, \omega) = \E_{p_{\rm data}(\rvx)}\E_{q_{\omega}(\rvz|\rvx)}[\nabla_{\beta_0}\log p_{\beta_0}(\rvx|\rvz)] 
\end{eqnarray}

\noindent \textbf{Learning prior model $\alpha$, $\beta_{>0}$:} For learning $\alpha_i$, the gradient is computed as
\begin{equation}
\begin{aligned}
-\nabla_{\alpha_i}L(\theta, \omega)  &= \E_{p_{\rm data}(\rvx)}\E_{q_{\omega}(\rvz|\rvx)}[\nabla_{\alpha_i}f_{\alpha_i}(\rvz_i)] \\
&-\E_{p_{\alpha, \beta_{>0}}(\rvz)}[\nabla_{\alpha_i}f_{\alpha_i}(\rvz_i)]
\end{aligned}
 \label{energy-grad-2}
\end{equation}
For learning $\beta_{>0}$, we compute the gradient as
\begin{equation}
\begin{aligned}
-\nabla_{\beta_i}L(\theta, \omega) &= \E_{p_{\rm data}(\rvx)q_{\omega}(\rvz|\rvx)}[\nabla_{\beta_i}\log p_{\beta_i}(\rvz_i|\rvz_{i+1})]\\
&- \E_{p_{\alpha, \beta_{>0}}(\rvz)}[\nabla_{\beta_i}\log p_{\beta_i}(\rvz_i|\rvz_{i+1})]
\end{aligned}
 \label{z-generator-grad-2}
\end{equation}
\noindent \textbf{Learning inference model $\omega$:} For learning $\omega_i$, the gradient is
\begin{equation}
\begin{aligned}
    -\nabla_{\omega_i}L(\theta, \omega) &= \nabla_{\omega_i}\E_{p_{\rm data}(\rvx)}[\E_{q_{\omega}(\rvz|\rvx)} \log p_{\beta_0}(\rvx|\rvz) \\
    &- D_{\KL}(q_{\omega}(\rvz|\rvx)||p_{\beta_{>0}}(\rvz))\\
    &+ \E_{q_\omega(\rvz|\rvx)}[\sum_{i=1}^{L} f_{\alpha_i}(\rvz_i)]]
\end{aligned}
 \label{q-generator-grad}
\end{equation}
We refer to detailed derivation in Appendix.\ref{sec:appendix-theorectical}.

\noindent \textbf{Divergence Perturbation.}
The KL joint minimization (Eqn.\ref{eq:kl_joint}) can be viewed as a surrogate of the MLE objective with the KL perturbation term,
\begin{eqnarray*}
&&D_{\mathrm{KL}}(q_{\omega}(\rvx, \rvz)||p_{\theta}(\rvx,\rvz)) \\
&=& D_{\mathrm{KL}}(p_{\rm data}(\rvx)||p_\theta(\rvx)) + D_{\mathrm{KL}}(q_\omega(\rvz|\rvx)||p_\theta(\rvz|\rvx))
\end{eqnarray*}
where the perturbation term $D_{\mathrm{KL}}(q_\omega(\rvz|\rvx)||p_\theta(\rvz|\rvx))$ measures the KL-divergence between inference distribution and generator posterior. The inference model is learned to directly match
the posterior distribution of the generator without expensive posterior sampling. In fact, such KL minimization in the joint space is closely related to evidence lower bound (ELBO) with the joint EBM as the prior model.

\section{Related Work}
\noindent \textbf{Hierarchical VAEs:} Variational auto-encoder (VAE) \cite{kingma2013auto} proposes variational learning by introducing an approximation of the true intractable posterior, which allows a tractable bound on log-likelihood to be formed. But the non-hierarchical structure can be limited in model expressivity and fails to capture different levels of abstraction. Hierarchical VAEs (HVAEs) \cite{NIPS2016_6ae07dcb,vahdat2020nvae,DBLP:conf/iclr/Child21,maaloe2019biva} consist of multiple layers of latent variables on top of the generator as a prior model, which can be used for learning complex data distribution and hierarchical representations. However, such models still focus on layer-wise relations while ignoring the intra-layer contextual relations at each layer.

\noindent \textbf{Energy-based models:} The energy-based models receive attention for being expressive and powerful in capturing contextual data regularities. The majority of existing works focus on the pixel space \cite{xiao2020vaebm,du2019implicit,du2020improved,yin2020analyzing,xie2022a,han2020joint,gao2020learning,Han_2019_CVPR}. Learning such EBMs can be done using MLE, where MCMC sampling is typically required in each learning iteration which can be computationally expensive upon data space. Instead, \cite{pang2020learning} proposes to build EBM on latent space where the energy function is considered as a correction of the non-informative Gaussian prior. The low dimensionality of the latent space makes EBM effective in capturing data regularities and can alleviate the burden of MCMC sampling. 

\noindent \textbf{Generator models with informative prior:} For generator models, the assumed Gaussian or uniform prior distribution can be non-informative and less expressive. To address this problem, recent works \cite{tomczak2018vae,ghosh2019variational,dai2019diagnosing,pang2020learning,aneja2021contrastive,xiao2022adaptive} propose to learn generator models with an informative prior, where RAE \cite{ghosh2019variational} constructs priors using rejection sampling, and Two-stage VAE \cite{dai2019diagnosing} propose to train an extra model for simple prior at the second stage to match the aggregated posterior distribution, while LEBM \cite{pang2020learning} and NCP-VAE \cite{aneja2021contrastive} instead learn EBMs on latent space to improve the expressivity of generator models. 

\section{Experiments}
To demonstrate the proposed method, we present extensive experiments, including (i) latent visualization, (ii) image synthesis, (iii) hierarchical representations, and (iv) analysis of latent space. To better understand the proposed model, we conduct various ablation studies based on the proposed EBM prior in Sec.\ref{section:ablation}. The parameter complexity is discussed in Sec.\ref{section:param}. \blfootnote{Our project page is available at \url{https://jcui1224.github.io/hierarchical-joint-ebm-proj}.}

\subsection{Latent Visualization}
\begin{figure}[h]
\centering
\includegraphics[width=.95\columnwidth]{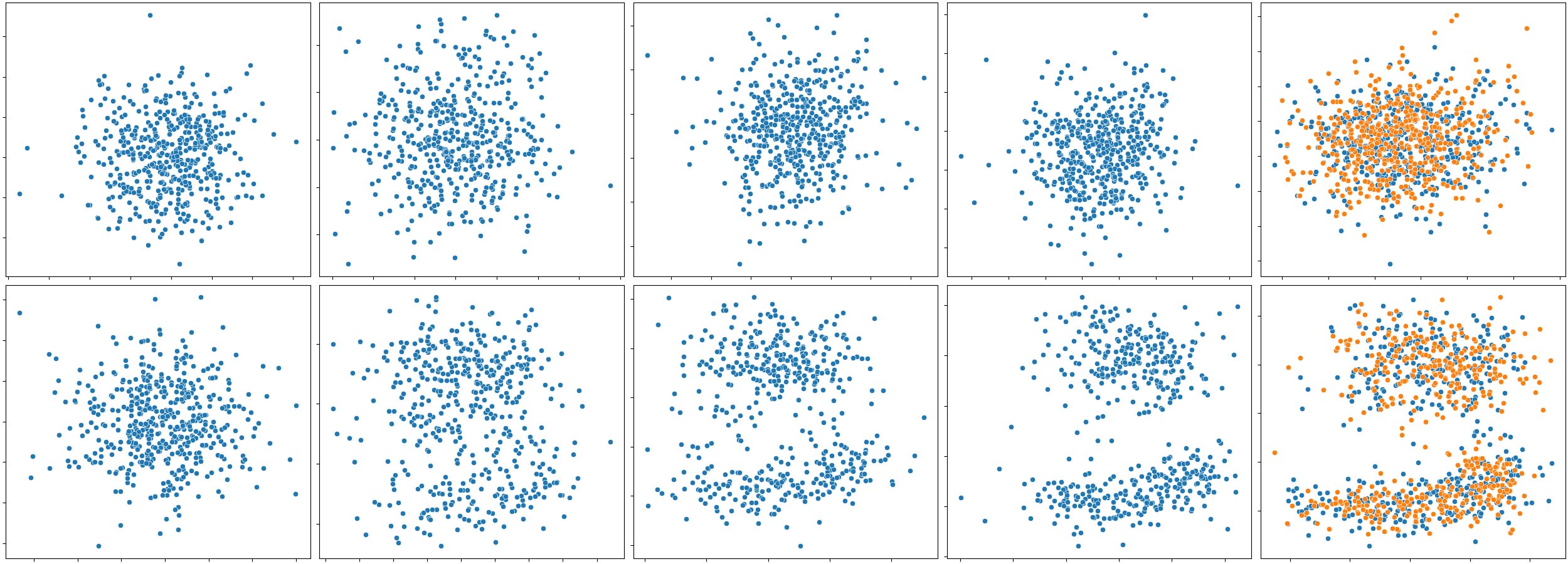}
\caption{Langevin transition on latent codes (bottom: $\rvz_1$, top: $\rvz_2$). \textbf{Blue} color indicates the transition of Langevin prior sampling. \textbf{Orange} color indicates latent codes inferred from inference model.}
\label{Fig.toy}
\end{figure}

 We examine the expressivity of our EBM prior model by latent visualization. We pick MNIST data with only digit classes `1' and `0' available, on which we train our 2-layer model with the latent dimension of each layer set to be 2 for better visualization. We train with $k=40$ steps for prior sampling and visualize the transition of Langevin dynamics on each layer for every 10 steps in Fig.\ref{Fig.toy}. It can be seen that the latent variables are first initialized from Gaussian noise and then can be tilted to match the multi-modal posterior, for which the standard Gaussian prior can be infeasible.

\begin{figure*}[t]
\centering
\includegraphics[width=0.95\textwidth]{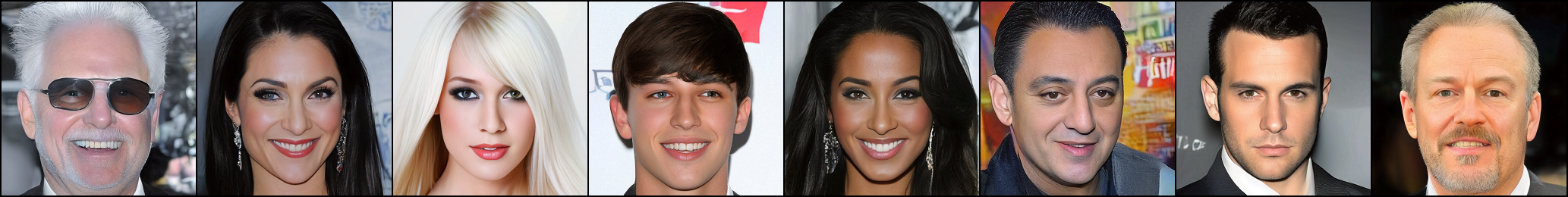}
\and\caption{Generated samples on CelebA-HQ-256 . FID = 9.89.}
\label{Fig.celeba256}
\end{figure*}

\subsection{Image Synthesis}
\label{sec-image-synthesis}
\noindent \textbf{Generator models with informative prior.} We evaluate the generation performance of the proposed joint model. If the model is well-trained, the multi-layer EBM prior model should render an expressive prior distribution leading to realistic synthesis. We benchmark our model against other generator models that assume standard Gaussian prior, such as VAE \cite{kingma2013auto}, Alternating Back-propagation (ABP) \cite{han2017alternating}, Ladder VAE (LVAE) \cite{NIPS2016_6ae07dcb}, and Short-run Inference (SRI) \cite{nijkamp2020learning}, as well as other generator models using informative prior, such as RAE \cite{ghosh2019variational}, Two-stages VAE (2s-VAE) \cite{dai2019diagnosing}, NCP-VAE \cite{aneja2021contrastive}, and LEBM \cite{pang2020learning}, where LEBM builds EBM for single layer latent variables, while ours contains a multi-layer structure. 

We train our model on SVHN \cite{37648}, CIFAR-10 \cite{krizhevsky2009learning} and CelebA-64 \cite{DBLP:journals/corr/LiuLWT14} and use Fr$\acute{e}$chet Inception Distance (FID) \cite{NIPS2017_8a1d6947} to quantitatively evaluate the generation quality. To make fair comparisons, we follow the standard protocol as in \cite{pang2020learning} and use the same generation model with convolutional structures. We use Langevin posterior sampling for the training, and the generation model is jointly learned (the result for variational learning is shown in Ablation Studies). The comparisons are shown in Tab.\ref{table.scratch-fid}, where the superior generation performance indicates the effectiveness of our model in learning a more expressive prior.

\begin{table}[h]
\centering
  \setlength{\belowcaptionskip}{-10pt}%
  \begin{tabular}{lccc}
    \toprule
    Model & SVHN & CelebA-64 & CIFAR-10 \\
    \midrule
    VAE \cite{kingma2013auto} & 46.78 & 65.75 & 106.37 \\
    LVAE (L=5) \cite{NIPS2016_6ae07dcb} & 39.26 & 53.40 & - \\
    ABP \cite{han2017alternating}& 49.71 & 51.50 & - \\
    SRI (L=5) \cite{nijkamp2020learning}& 35.32 & 47.95 & - \\
    \midrule
    RAE \cite{ghosh2019variational}& 42.02 & 40.95 & 74.16 \\
    2s-VAE \cite{dai2019diagnosing}& 42.81 & 44.40 & 72.90\\
    NCP-VAE \cite{aneja2021contrastive}& 33.23 & 42.07 & 78.06\\
    LEBM \cite{pang2020learning}& 29.44 & 37.87 & 70.15\\
    \bottomrule
    Ours (L=2) & \textbf{26.81} & \textbf{33.60} & \textbf{66.32}\\
    \bottomrule
  \end{tabular}
\caption{FID$(\downarrow)$ for our model and baselines on SVHN, CelebA (64 x 64), and CIFAR-10.}
\label{table.scratch-fid}
\end{table}

\noindent \textbf{Toward deep hierarchical models.} We then consider the modern deep hierarchical structures as our multi-layer generator and explore the potential of the joint EBM prior for better generation. We adopt the two-stage training \cite{xiao2020vaebm, aneja2021contrastive} where the deep multi-layer generator $p_{\beta_{>0}}(\rvz)$ and inference model $q_\omega(\rvz|\rvx)$ are trained in the first stage by maximizing the ELBO as in VAEs, with the pre-trained models, our joint EBM prior model can then be learned in the second stage where the posterior samples are directly obtained from the pre-trained inference model $q_\omega(\rvz|\rvx)$ and prior samples can be obtained via Langevin sampling with change of variable on the generator $p_{\beta>0}(\rvz)$ (see details in Appendix.\ref{sec:appendix-cov}). 

\begin{figure}[t]
\centering
\includegraphics[width=0.85\columnwidth,height=2cm]{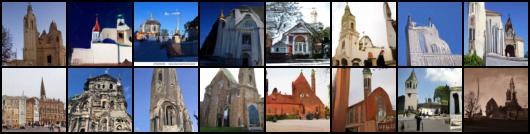}
\caption{Generated samples on LSUN-Church-64. FID = 8.38.}
\label{Fig.church64}
\end{figure}

\begin{figure}[t]
\centering
\includegraphics[width=0.85\columnwidth,height=2cm]{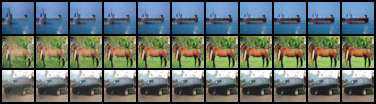}
\caption{Langevin transitions on CIFAR-10. FID = 11.34.}
\label{Fig.cifar-gen}
\end{figure}

We consider NVAE \cite{vahdat2020nvae}, a modern hierarchical VAE, for the first stage training, and we train our joint EBM prior in the second stage. For prior sampling in the second stage training, we employ similar \textit{reparametrized sampling} scheme as in \cite{xiao2020vaebm} via provided code\footnote{https://github.com/NVlabs/VAEBM} in order to better traverse the deep hierarchical latent space with different scales. We examine our model on CIFAR-10, CelebA-HQ-256 \cite{karras2017progressive}, and LSUN-Church-64 \cite{yu15lsun}. The qualitative results for CelebA-HQ-256 and LSUN-Church-64 are shown in Fig.\ref{Fig.celeba256} and Fig.\ref{Fig.church64}. For CelebA-HQ-256, we synthesize with adjusted batch-normalization as used in \cite{aneja2021contrastive,xiao2020vaebm}. We also visualize the Langevin transition on CIFAR-10 in Fig.\ref{Fig.cifar-gen} where the quality of synthesis improves as the Langevin progresses. We refer to more results in Appendix.\ref{sec:appendix-additional}

The quantitative results are shown in Tab.\ref{table.cifar-backbone-fid} and Tab.\ref{table.celeba256-backbone-fid}. We consider the baseline models, including NCP-VAE \cite{aneja2021contrastive} and VAEBM \cite{xiao2020vaebm}, which also recruit NVAE as their backbone model, and other powerful deep generative models, such as GANs \cite{brock2018large,karras2020training}, score-based models \cite{song2019generative,ho2020denoising} and EBMs \cite{du2019implicit,du2020improved,Han_2019_CVPR,yin2020analyzing} on data space. Compared to NVAE backbone model, our joint EBM prior model can significantly improve the fidelity of generated samples while only accounting for negligible overhead (see Parameter Efficiency in Sec.\ref{section:param}). In comparison with other powerful deep generative models, we also achieve competitive generation performance.

\begin{table}[t]
\centering
  \setlength{\belowcaptionskip}{-10pt}%
\resizebox{0.75\columnwidth}{!}{
  \begin{tabular}{lcc}
\toprule
    Method  & IS & FID\\
\toprule
    NVAE$^*$\cite{vahdat2020nvae} & 5.30 & 37.73 \\
    Ours & 8.99 &11.34 \\
\cmidrule(lr){0-0}
    NCP-VAE\cite{aneja2021contrastive} & - &24.08\\
    VAEBM\cite{xiao2020vaebm} & 8.43 &12.19 \\
\midrule
\textbf{Other EBMs} & &\\
    IGEBM\cite{du2019implicit} & 6.78 &38.2 \\
    ImprovedCD\cite{du2020improved} & 7.85 &25.1 \\
    Divergence Triangle\cite{Han_2019_CVPR} & - &30.10 \\
    Adv-EBM\cite{yin2020analyzing} & 9.10 & 13.21 \\
\toprule
    \textbf{Other Likelihood Models} && \\
    GLOW\cite{kingma2018glow} & 3.92 & 48.9 \\
    PixelCNN\cite{van2016pixel} & 4.60 &65.93\\
\toprule
    \textbf{GANs}+\textbf{Score-based Models} && \\
    BigGAN\cite{brock2018large} & 9.22 & 14.73 \\
    StyleGANv2 w/o ADA\cite{karras2020training} & 8.99 & 9.9 \\
    NCSN\cite{song2019generative} & 8.87 & 25.32 \\
    DDPM\cite{ho2020denoising} & 9.46 & 3.17\\
\toprule
  \end{tabular}
}
\caption{IS$(\uparrow)$ and FID$(\downarrow)$ for our model and baselines on CIFAR-10. Model$^*$ indicates our backbone model.}
\label{table.cifar-backbone-fid}
\end{table}

\begin{table}[t]
\centering
  \setlength{\belowcaptionskip}{-10pt}%
\resizebox{0.95\columnwidth}{!}{
  \begin{tabular}{l c c}
\toprule
    Model  & CelebA-HQ-256 & LSUN-Church-64\\
\toprule
    NVAE$^*$ \cite{vahdat2020nvae} & 30.25 & 38.13\\
    Ours & 9.89 & 8.38\\
\cmidrule(lr){0-0}
    NCP-VAE\cite{aneja2021contrastive} & 24.79 & -\\
    VAEBM\cite{xiao2020vaebm} & 20.38 & 13.51\\
\toprule
    Adv-EBM\cite{yin2020analyzing} & 17.31 & 10.84\\
    GLOW\cite{kingma2018glow} & 68.93 & 59.35\\
    PGGAN\cite{karras2017progressive} & 8.03 & 6.42\\
\bottomrule
  \end{tabular}
}
\caption{FID$(\downarrow)$ for our model and baselines on CelebA-HQ-256 and LSUN-Church-64. Model$^*$ indicates backbone model.}
\label{table.celeba256-backbone-fid}
\end{table}

\begin{figure*}[t]
\centering
\includegraphics[width=0.95\textwidth]{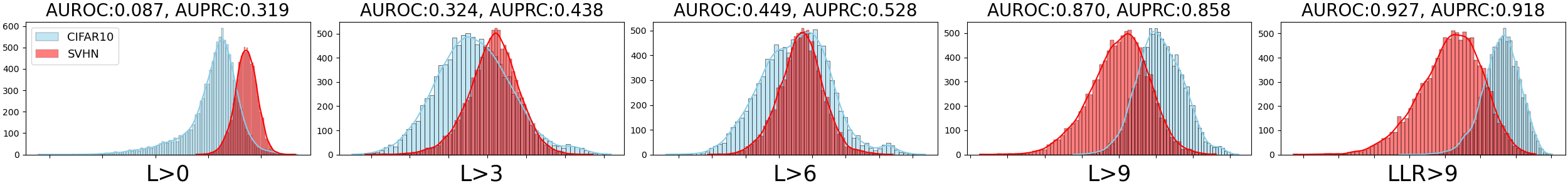}
\caption{Histograms of density of $L_{\text{EBM}}^{>k}$ with AUROC$(\uparrow)$ and AUPRC$(\uparrow)$ for CIFAR-10 (in) / SVHN (out).}
\label{Fig.ood_density}
\end{figure*}

\subsection{Hierarchical Representations}

\noindent\textbf{Hierarchical sampling.} To examine our model in learning hierarchical representation, we employ hierarchical sampling to illustrate the learned representation at different layers. In particular, we first sample one group of latent vectors from EBM prior and hold them as fixed constants, then we randomly sample multiple groups of latent vectors to replace the fixed latent vectors at different layers. This allows us to visualize the variation in representation across layers.

\begin{figure}[h]
\centering
\includegraphics[width=0.325\columnwidth]{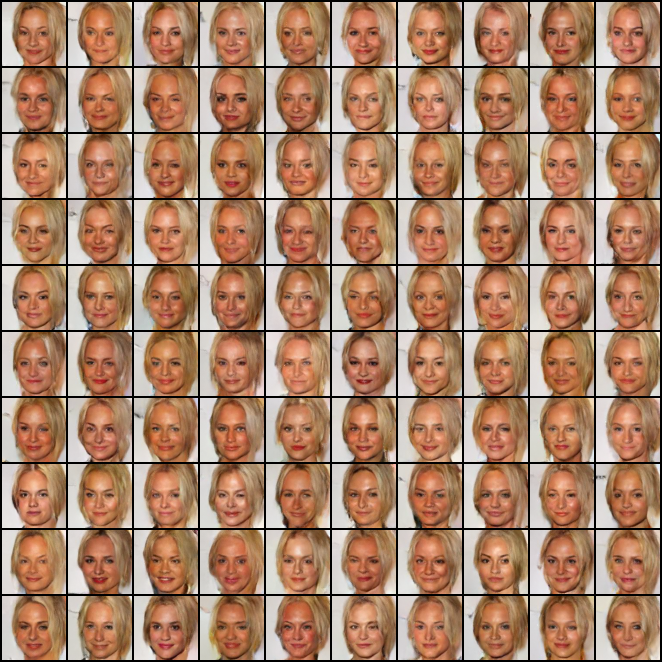}
\includegraphics[width=0.325\columnwidth]{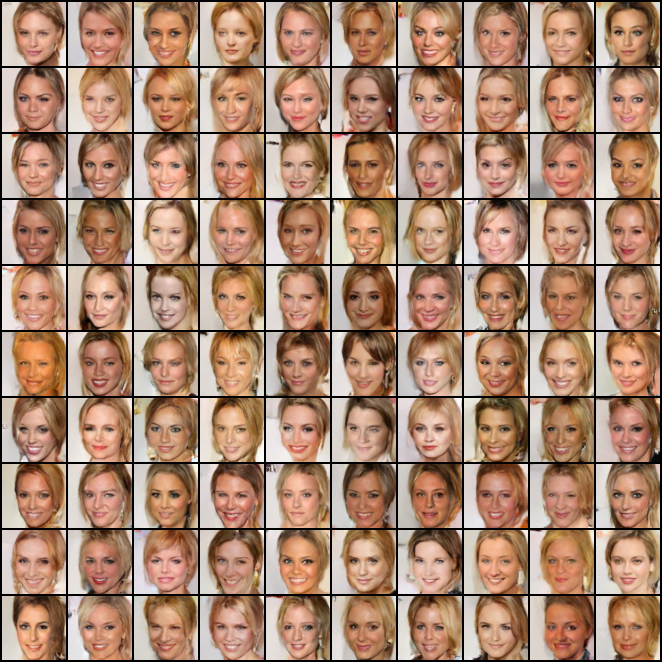}
\includegraphics[width=0.325\columnwidth]{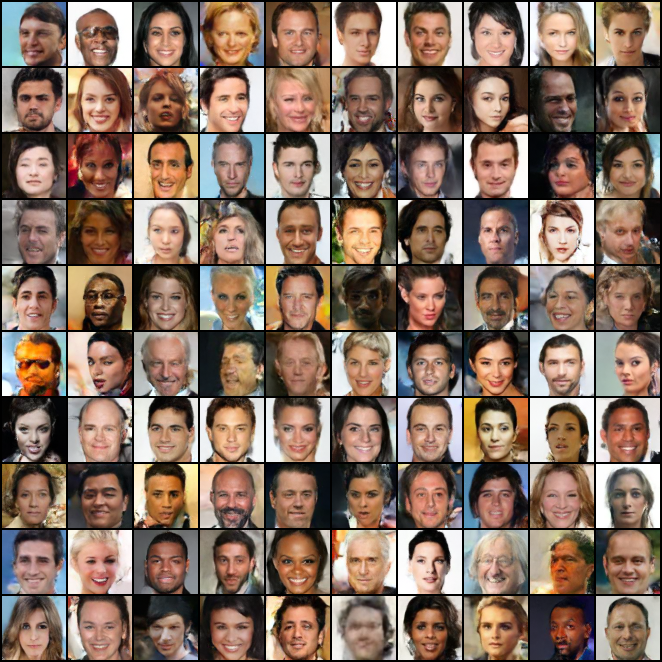}
\includegraphics[width=0.325\columnwidth]{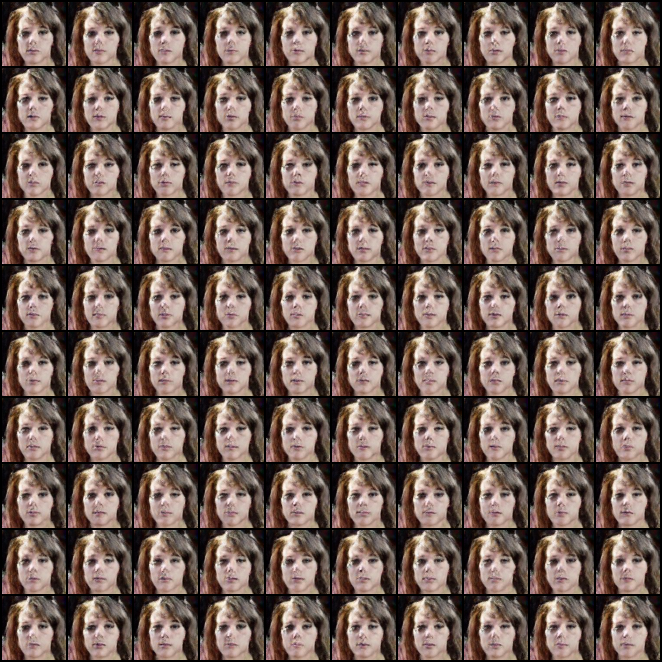}
\includegraphics[width=0.325\columnwidth]{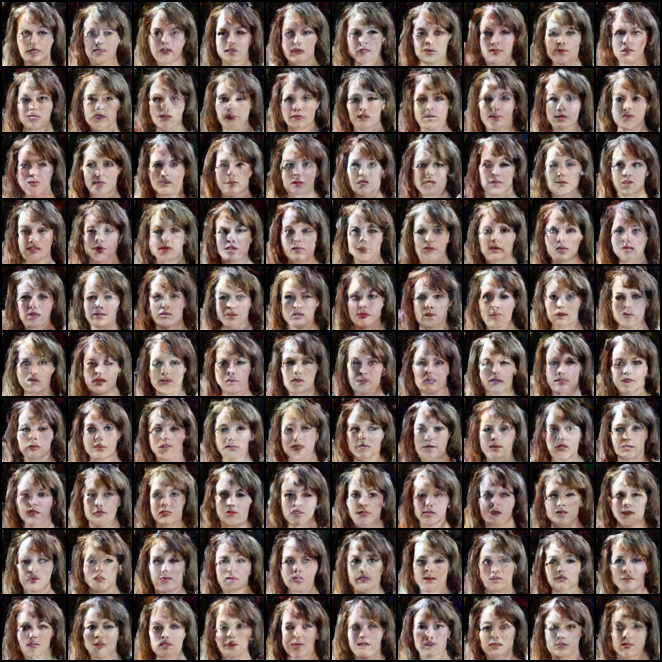}
\includegraphics[width=0.325\columnwidth]{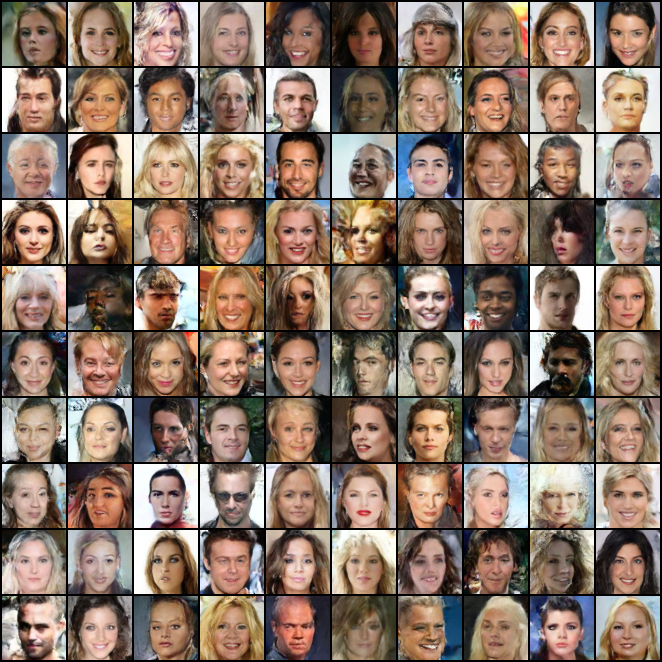}
\and\caption{Hierarchical sampling for Gaussian prior model (\textbf{bottom}) and EBM prior model (\textbf{top}). From \textbf{left panel} to \textbf{right panel}, latent vectors are sampled from the bottom layers to the top layers. Detailed sampling method can be referred in Figure 3 in \cite{zhao2017learning}.}
\label{Fig.hs_ebm}
\end{figure}

We apply our joint EBM prior to BIVA\cite{maaloe2019biva} on CelebA-64. For training, we reuse the two-stage training scheme, where we recruit BIVA\footnote{https://github.com/vlievin/biva-pytorch} for the first stage training and train our EBM prior model in the second stage by using the \textit{reparametrized sampling} \cite{xiao2020vaebm} method (similar as deep hierarchical models in Sec.\ref{sec-image-synthesis}). We show the results of hierarchical sampling in Fig.\ref{Fig.hs_ebm} and observe that BIVA presents minor changes at the bottom and middle layers, while the proposed joint EBM prior model can show variations of different levels. Note that it is a challenging task for conditional hierarchical models \cite{zhao2017learning}, on which the improvement thus suggests that the proposed method is capable of learning hierarchical representations for multi-layer generator models. Additional results are referred to Appendix.\ref{sec:appendix-hierarchical}.

\noindent\textbf{Out-of-distribution detection.} Next, we conduct out-of-distribution (OOD) detection to further evaluate the hierarchical representations. Typically, low-level representations (e.g., edges, corners) can be shared across data which in turn leads to high-confidence reconstructions for OOD examples, while high-level semantic ones have fewer correlations across different data and shall be more discriminative for OOD detection. Inspired by \cite{havtorn2021hierarchical}, we consider an unnormalized log-posterior as the decision function for EBM prior model, which is defined as

\begin{eqnarray}
\label{ood-L>K-ebm}
L^{>k}_{\text{EBM}} &=&\E_{\rvz_{>k} \sim q_{\omega}(\rvz|\rvx),\rvz_{\le k} \sim p_{\beta_{>0}, \alpha}(\rvz)}[\log p_{\beta_0}(\rvx|\rvz)  \nonumber\\
 &+& \log p_{\beta_{>0}}(\rvz) + \sum_{i=1}^{L}f_{\alpha_i}(\rvz_i)]
\end{eqnarray}
where latent codes above the $k$-th layer are inferred from inference model and kept fixed, and those below the $k$-th layer are sampled from EBM prior via the \textit{reparametrized sampling}\footnote{https://github.com/NVlabs/VAEBM} \cite{xiao2020vaebm} with fixed inferred latent codes. With $k=0$, all layers of latent vectors are inferred from $q_\omega(\rvz|\rvx)$. With a higher value of $k$, less inferred low-level representations are used, which should render better performance in OOD detection. In addition, we can also compute a subtraction between $L^{>0}_{\text{EBM}}$ and $L^{>k}_{\text{EBM}}$ as a surrogate of the likelihood-ratio which is shown to be effective for OOD detection \cite{havtorn2021hierarchical}. We compute the subtraction as
\begin{eqnarray}
\label{ood-LLR>K-ebm}
LLR^{>k}_{\text{EBM}} &=& L^{>0}_{\text{EBM}} - L^{>k}_{\text{EBM}}
\end{eqnarray}

We follow standard protocols and apply our EBM prior model with BIVA on CIFAR-10 and use SVHN as OOD data for testing. In Fig.\ref{Fig.ood_density}, we show the density of in-distribution and OOD data by computing the unnormalized log-posterior with increased $k$, and we use AUROC, AUPRC to quantitatively evaluate the performance. It can be seen that as $k$ increases, relatively lower log-likelihoods are assigned to OOD data, which in turn renders better detection performance (higher AUROC and AUPRC). More importantly, we observe that the backbone model BIVA achieves the best detection performance of 0.885 for AUROC, while our models achieve 0.927 with the adapted decision function. This further verifies that the hierarchical representations can be learned within our multi-layer structure.

\subsection{Analysis of Latent Space}
\label{section:ebm-exp}
\noindent \textbf{Long-run langevin transition.} In this section, we examine the energy landscape of our joint EBM prior model. If the EBM is well-learned, the energy prior should naturally render local modes of the energy function, and traversing these local modes should present realistic synthesized examples and steady-state energy scores. Existing EBMs typically have oversaturated images via long-run Langevin dynamics as observed in \cite{nijkamp2020anatomy}. Training an EBM that learns steady-state energy scores over realistic images can be useful but challenging. 

\begin{figure}[h]
\centering
\includegraphics[width=0.95\columnwidth]{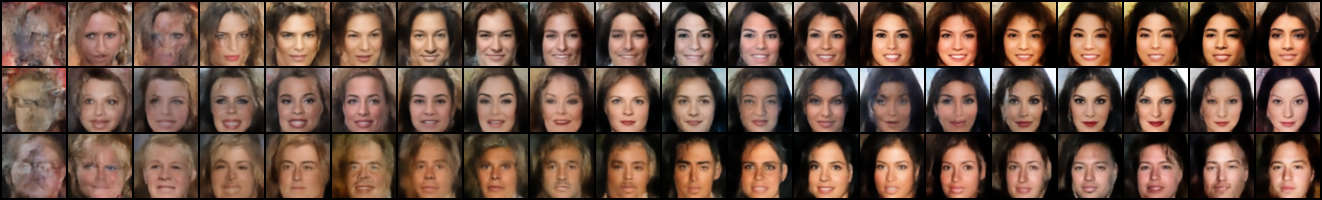}
\includegraphics[width=0.95\columnwidth]{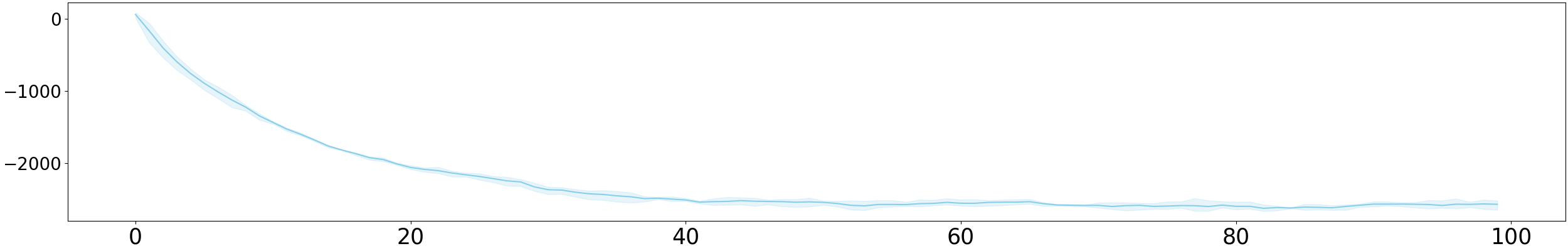}
\includegraphics[width=0.95\columnwidth]{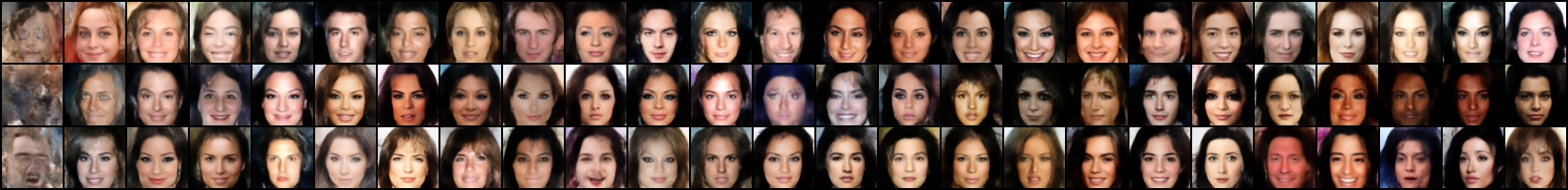}
\includegraphics[width=0.95\columnwidth]{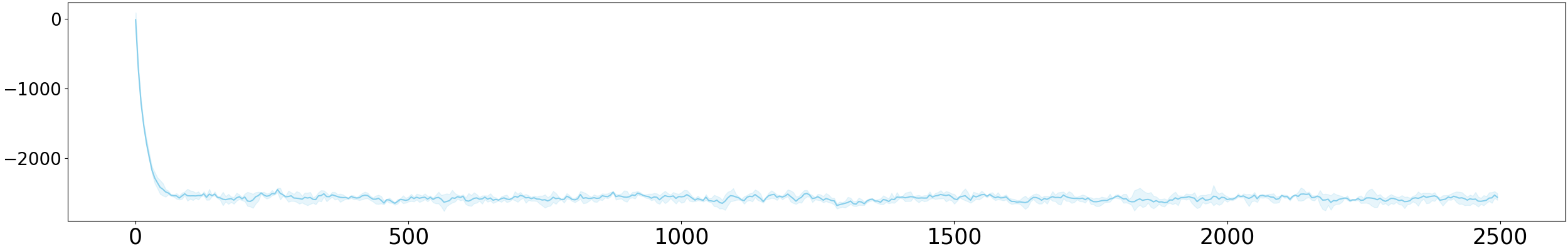}
\caption{Trajectory in data space and energy profile in Langevin transition. \textbf{Top:} Langevin transition with 100 steps. \textbf{Bottom:} Langevin transition with 2500 steps.}
\label{Fig.short-run}
\end{figure}

We train our model on CelebA-64 using Langevin dynamic for 40 steps. We then run 100 and 2500 Langevin steps to examine the learned energy landscape. We show the synthesis and corresponding energy profile in Fig.\ref{Fig.short-run}. 
It can be seen that generated examples become sharper for the first 40 steps as it starts from the referenced distribution $p_{\beta_{>0}}(\rvz)$ toward the learned energy prior $p_{\beta_{>0},\alpha}(\rvz)$, and the energy fluctuates around some constant. For long-run 2500 steps, it is worth noting that our EBM prior model delivers diverse and realistic synthesis, and it does not exhibit the oversaturated phenomenon. This suggests that the learned EBM could mix well between different local modes of the learned energy prior.

\noindent \textbf{Anomaly Detection.} We further evaluate how our joint EBM prior model could benefit the anomaly detection (AD) task. Different from OOD detection, AD requires one class (e.g., one-digit class from MNIST) of data to be held out as anomaly for training, and both normal (e.g., other nine-digit classes from MNIST) and anomalous data are used for testing. 

The proposed prior model is built on the joint of all layers of latent variables. If it is well learned, the posterior $q_\omega(\rvz|\rvx)$ could form a discriminative joint latent space that has separated probability densities for normal and anomalous data. We use un-normalized log-posterior $L^{>0}_{\text{EBM}}$ as our decision function and train our model on MNIST with each class held out as an anomalous class. We consider the baseline models that also adopt an inferential mechanism, such as VAE \cite{kingma2013auto}, MEG \cite{kumar2019maximum}, BiGAN-$\sigma$ \cite{zenati2018efficient}, OT-SRI \cite{An_2021_CVPR}, and LEBM \cite{pang2020learning} which assumes single-layer latent space and is closely related to our method. Tab.\ref{table.ad} shows the results of AUPRC scores averaged over the last 10 epochs to account for the variance. To make fair comparisons, we follow the protocols in \cite{pang2020learning,kumar2019maximum,zenati2018efficient,An_2021_CVPR}.


\begin{table}[h]
\centering
\resizebox{0.99\columnwidth}{!}{
\begin{tabular}{c | c | c | c | c | c } 
 \hline
 Heldout Digit & 1 & 4 & 5 & 7 & 9 \\ 
 \hline
 VAE\cite{kingma2013auto} & 0.063 & 0.337 & 0.325 & 0.148 & 0.104 \\
 MEG\cite{kumar2019maximum} & 0.281 $\pm$ 0.035  & 0.401 $\pm$ 0.061  & 0.402 $\pm$ 0.062  & 0.290 $\pm$ 0.040  & 0.342 $\pm$ 0.034 \\ 
 BiGAN-$\sigma$\cite{zenati2018efficient} & 0.287 $\pm$ 0.023  & 0.443 $\pm$ 0.029  & 0.514 $\pm$ 0.029  & 0.347 $\pm$ 0.017  & 0.307 $\pm$ 0.028 \\ 
  OT-SRI\cite{An_2021_CVPR} & 0.353 $\pm$ 0.021  & 0.770 $\pm$ 0.024  & 0.726 $\pm$ 0.030  & 0.550 $\pm$ 0.013  & 0.555 $\pm$ 0.023 \\
 LEBM\cite{pang2020learning} & 0.336 $\pm$ 0.008  & 0.630 $\pm$ 0.017  & 0.619 $\pm$ 0.013  & 0.463 $\pm$ 0.009  & 0.413 $\pm$ 0.010 \\
 Ours & \textbf{0.470 $\pm$ 0.009}  & \textbf{0.941 $\pm$ 0.001}  & \textbf{0.964 $\pm$ 0.003}  & \textbf{0.815 $\pm$ 0.004}  & \textbf{0.796 $\pm$ 0.004} \\
 \hline
\end{tabular}%
}
\caption{AUPRC scores for unsupervised anomaly detection.}
\label{table.ad}
\end{table}

\subsection{Ablation Studies.}
\label{section:ablation}

\noindent\textbf{Informative prior vs. complex generator:} We examine the expressivity endowed with the joint EBM prior by comparing it to hierarchical Gaussian prior model. We use the same experimental setting as reported in Tab.5 in main text and increase the complexity of generator model for hierarchical Gaussian prior. The FID results are shown in Tab.\ref{table.gaussian-gen}, in which the Gaussian prior models exhibit an improvement in performance as the generator complexity increases. However, even with eight times more parameters, hierarchical Gaussian prior models still have an inferior performance compared to our joint EBM prior model.
\begin{table}[h]
\centering
\resizebox{0.99\columnwidth}{!}{
  \begin{tabular}{l|cccc}
    \toprule
    Ours & same generator & 2x parameters & 4x parameters & 8x parameters \\
    \midrule
    \textbf{28.60} & 42.03 & 39.82 & 37.75 & 36.10\\
    \bottomrule
  \end{tabular}
}
\caption{Comparison on Gaussian prior and our EBM prior.}
\label{table.gaussian-gen}
\end{table}

\noindent \textbf{Complexity of EBM.} The energy function $f_{\alpha_i}(\rvz_i)$ is parameterized by a small multi-layer perceptron. To better understand the effectiveness of our EBM, we fix the generator network $p_{\beta_0}(\rvz|\rvx)$ and increase hidden units (\textbf{nef}) of energy functions. We train our model on CIFAR-10 with \textbf{nef} increasing from 10 to 100. The results are shown in Tab.\ref{table.nef}. The larger capacity of the EBM could in general render better model performance.
\begin{table}[h]
\centering
\resizebox{0.85\columnwidth}{!}{
\begin{tabular}{l | c c c c } 
\toprule
nef & nef = 10 & nef = 20 & nef = 50 & nef = 100\\
\midrule
FID & 69.73 & 68.45 & 67.88 & \textbf{66.32}\\
\bottomrule
\end{tabular}
}
\caption{FID for increasing hidden units (\textbf{nef}) of EBM}
\label{table.nef}
\end{table}

\noindent \textbf{MCMC sampling vs. Inference model.} Two posterior sampling schemes using MCMC and inference model are compared in Tab.\ref{table.mcmc_vs_inf} in terms of FID and wall-clock training time (per-iteration). 
The MCMC posterior sampling renders better FID as it is more accurate in inference \cite{nijkamp2020learning,han2017alternating}, but it can be computationally heavy. While the inference model is efficient in learning but can be less accurate. For deep hierarchical structures, the variational learning with inference model is preferred due to its efficiency. 

\begin{table}[h]
\centering
\setlength{\belowcaptionskip}{-10pt}%
\resizebox{0.85\columnwidth}{!}{
\begin{tabular}{l | c c c} 
\toprule
MCMC / Inf & SVHN & CelebA-64 & CIFAR-10\\
\midrule
FID        & 26.81 / 28.60 & 33.60 / 36.12 & 66.32 / 68.45\\
Time(s)   & 0.478 / 0.232 & 0.920 / 0.246 & 0.568 / 0.256\\
\bottomrule
\end{tabular}
}
\caption{FID and training time for MCMC posterior sampling and variational learning.}
\label{table.mcmc_vs_inf}
\end{table}

\noindent \textbf{Langevin steps.} We explore the different number of Langevin steps in prior sampling for training on CIFAR-10. The results of FID and corresponding training time are shown in Tab.\ref{table.steps_in_prior}. We observe that the Langevin step $k$ increasing from 10 to 40 can improve the generation quality, while for steps more than 40, it only has minor impacts on the improvement but with increased training overhead. We thus report the result of $k=40$ in Tab.\ref{table.scratch-fid}.

\begin{table}[h]
\centering
\setlength{\belowcaptionskip}{-10pt}%
\resizebox{0.85\columnwidth}{!}{
\begin{tabular}{l | c c c c c} 
\toprule
steps $k$ & $k$ = 10 & $k$ = 20 & $k$ = 40 & $k$ = 80 & $k$ = 100\\
\midrule
FID     & 69.42 & 67.58 & \textbf{66.32} & 66.03 & 65.86\\
Time(s) & 0.312 & 0.480 & 0.568 & 0.741 & 0.837\\
\bottomrule
\end{tabular}
}
\caption{FID and training time for increasing MCMC steps in prior sampling.}
\label{table.steps_in_prior}
\end{table}

\begin{figure}[t]
\centering
\setlength{\belowcaptionskip}{-10pt}%
\includegraphics[width=0.42\columnwidth]{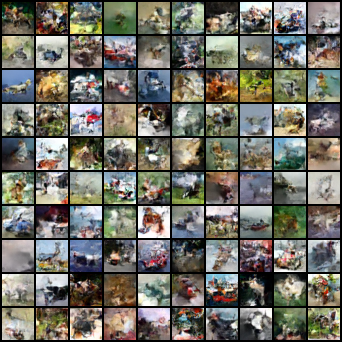}
\includegraphics[width=0.42\columnwidth]{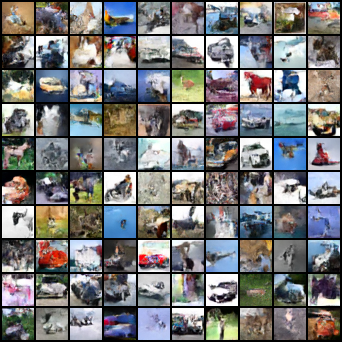}
\and\caption{Generated images on CIFAR-10. \textbf{Left:} HVAE. FID = 79.57 \textbf{Right:} Ours. FID = 49.50}
\label{Fig.cifar_hvae}
\end{figure}

\begin{figure}[h]
\centering
\setlength{\belowcaptionskip}{-8pt}%
\includegraphics[width=0.42\columnwidth]{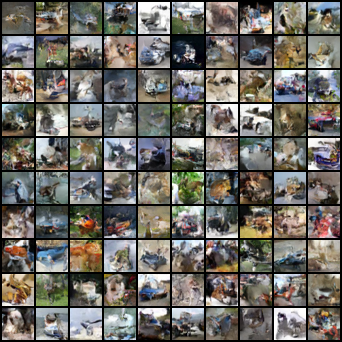}
\includegraphics[width=0.42\columnwidth]{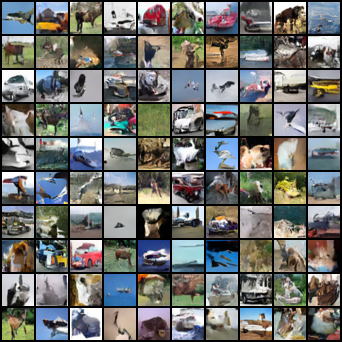}
\and\caption{Generated images on CIFAR-10. \textbf{Left:} BIVA. FID = 66.37 \textbf{Right:} Ours. FID = 25.87}
\label{Fig.cifar_biva}
\end{figure}

\noindent \textbf{Other backbone models:} We also examine the generation performance of our joint EBM prior on other multi-layer generator models, such as BIVA and HVAE. We implement the HVAE and BIVA using the provided codes\footnote{https://github.com/JakobHavtorn/hvae-oodd}\footnote{https://github.com/vlievin/biva-pytorch}. We show the image synthesis and corresponding FID scores in Fig.\ref{Fig.cifar_hvae} and Fig.\ref{Fig.cifar_biva}. It can be seen that the proposed method is expressive in generating sharp image synthesis and can be applied to different multi-layer generator models. 





\subsection{Parameter Efficiency}
\label{section:param}
It is crucial to analyze the parameter complexity when comparing the generation performance. In Tab.\ref{table.scratch-fid}, we build our model with two layers of latent variables on top of the generator used in \cite{pang2020learning}. The additional layer accounts for only 1\% overhead in total parameter complexity compared to LEBM \cite{pang2020learning}. For deep hierarchical models, we apply our joint EBM prior model on latent space which brings minimum overhead. The parameter complexity of the backbone NVAE and our EBM model is shown in Tab.\ref{table.param-complexity}.

\begin{table}[h]
\centering
\setlength{\belowcaptionskip}{-10pt}%
\resizebox{0.99\columnwidth}{!}{
\begin{tabular}{l | c c c} 
\toprule
NVAE / EBM & CIFAR-10 & CelebA-HQ-256 & LSUN-Church-64\\
\midrule
FID & 39.73 / \textbf{11.34} & 30.25 / \textbf{9.89} & 38.13 / \textbf{8.38}\\
Parameters & 257M / 9M (3\%) & 375M / 18M (4\%) & 65M / 5M (7\%)\\
\bottomrule
\end{tabular}
}
\caption{FID and parameter complexity for backbone model and EBM.}
\label{table.param-complexity}
\end{table}

\noindent \textbf{NVAE with Gaussian decoder:} In addition, we also consider NVAEs with a Gaussian decoder. Note that the discrete logistic decoder aims to conditionally models the pixels of images between different channels, while Gaussian decoder is a statistical simple model that predicts pixels independently. We use the NVAE that has 30 groups on CIFAR-10 and 20 groups on CelebA-HQ-256 as used in \cite{aneja2021contrastive,xiao2020vaebm}. The results of FID and parameter complexity are shown in Tab.\ref{table.gaussian-nvaes}, where our EBM prior still can largely improve the generation performance while only accounting for very small overhead in parameter complexity.

\begin{table}[h]
\centering
\setlength{\belowcaptionskip}{-10pt}%
\resizebox{0.99\columnwidth}{!}{
\begin{tabular}{l | c c c} 
\toprule
NVAE / EBM  & FID & Parameters & NVAE Group\\
\midrule
CIFAR10  & 52.45 / 14.92 & 130M / 10M (7.6\%) & 30\\
CelebA HQ 256  & 46.32 / 22.86 & 365M / 9M (2.4\%)& 20\\
\bottomrule
\end{tabular}
}
\caption{Parameter complexity and FID results based on NVAE with Gaussian decoder.}
\label{table.gaussian-nvaes}
\end{table}

\section{Conclusion}
we propose a joint EBM prior for multi-layer generator models, which can effectively capture the intra-layer relations at each layer and jointly correct the latent variables from all layers. We present a joint training scheme via MLE and further develop a variational learning scheme for efficient inference. Our comprehensive experiments demonstrate the effectiveness of the proposed method. 

{\small
\bibliographystyle{ieee_fullname}
\bibliography{egbib}
}

\clearpage
\begin{alphasection}
\section{Theoretical Derivations}\label{sec:appendix-theorectical}
\subsection{Maximum Likelihood Estimation}\label{sec:appendix-mle}
Recall that $ \nabla_{\theta}\log p_{\theta}(\rvx) = \E_{p_{\theta}(\rvz|\rvx)}[\nabla_{\theta}\log p_{\beta_0}(\rvx|\rvz)] + \E_{p_{\theta}(\rvz|\rvx)}[\nabla_{\theta}\log p_{\alpha,\beta_{>0}}(\rvz)]$, where $\theta=(\alpha,\beta_0,\beta_{>0})$. For the learning gradient of prior model $(\alpha_i,\beta_{>0})$, we compute $\E_{p_{\theta}(\rvz|\rvx)}[\nabla_{\alpha_i,\beta_{>0}}\log p_{\alpha,\beta_{>0}}(\rvz)]$ as
\begin{eqnarray}
 \label{normalization-grad-alpha}
     &&\nabla_{\alpha_i}\log p_{\theta}(\rvx)  = \E_{p_{\theta}(\rvz|\rvx)}[\nabla_{\alpha_i}\log p_{\alpha, \beta_{>0}}(\rvz)]\\
     &=& \E_{p_{\theta}(\rvz|\rvx)}[\nabla_{\alpha_i}f_{\alpha_i}(\rvz_i)] - \nabla_{\alpha_i}\log \mathrm{Z}_{\alpha, \beta_{>0}}\nonumber \\
     \nonumber\\
 \label{normalization-grad-beta}
      &&\nabla_{\beta_i}\log p_{\theta}(\rvx)  = \E_{p_{\theta}(\rvz|\rvx)}[\nabla_{\beta_i}\log p_{\alpha, \beta_{>0}}(\rvz)]\\
      &=& \E_{p_{\theta}(\rvz|\rvx)}[\nabla_{\beta_i}\log p_{\beta_i}(\rvz_i|\rvz_{i+1})] - \nabla_{\beta_i}\log \mathrm{Z}_{\alpha, \beta_{>0}}\nonumber
\end{eqnarray}
where $\mathrm{Z}_{\alpha, \beta_{>0}} = \int \exp{[f_{\alpha}(\rvz)]}p_{\beta_{>0}}(\rvz) d\rvz$. Therefore, for $\nabla_{\alpha_i} \log \mathrm{Z}_{\alpha, \beta_{>0}}$, we have
\begin{eqnarray}
 \label{normalization-grad-2}
     &&\nabla_{\alpha_i}\log \mathrm{Z}_{\alpha, \beta_{>0}}\\
     &=& \frac{1}{\mathrm{Z}_{\alpha, \beta_{>0}}} \int\nabla_{\alpha_i}\exp{[\sum_{i=1}^{L}f_{\alpha_i}(\rvz_i)]}p_{\beta_{>0}}(\rvz) d\rvz \nonumber \\
     &=& \int p_{\alpha, \beta_{>0}}(\rvz) \nabla_{\alpha_i}f_{\alpha_i}(\rvz_i) d\rvz \nonumber \\
     &=& \E_{p_{\alpha, \beta_{>0}}(\rvz)} [\nabla_{\alpha_i}f_{\alpha_i}(\rvz_i)] \nonumber
\end{eqnarray}
For $\nabla_{\beta_{>0}} \log \mathrm{Z}_{\alpha, \beta_{>0}}$, we have
\begin{eqnarray}
 \label{normalization-grad-3}
     &&\nabla_{\beta_i}\log \mathrm{Z}_{\alpha, \beta_{>0}}\\
     &=& \frac{1}{\mathrm{Z}_{\alpha, \beta_{>0}}} \int\exp{[f_{\alpha}(\rvz)]}\nabla_{\beta_i}\prod_{i=1}^{L-1}p_{\beta_i}(\rvz_i|\rvz_{i+1})p(\rvz_{L})  d\rvz \nonumber \\
     &=& \int p_{\alpha, \beta_{>0}}(\rvz) \nabla_{\beta_i}\log p_{\beta_i}(\rvz_i|\rvz_{i+1}) d\rvz \nonumber \\
     &=& \E_{p_{\alpha, \beta_{>0}}(\rvz)} [\nabla_{\beta_i}\log p_{\beta_i}(\rvz_i|\rvz_{i+1})] \nonumber
\end{eqnarray}
By applying Eqn.\ref{normalization-grad-2} to Eqn.\ref{normalization-grad-alpha}, we have
\begin{eqnarray}
 \label{energy-grad-appendix}
\nabla_{\alpha_i}\log p_{\theta}(\rvx)  &=& \E_{p_{\theta}(\rvz|\rvx)}[\nabla_{\alpha_i}f_{\alpha_i}(\rvz_i)] \\
&-&\E_{p_{\alpha, \beta_{>0}}(\rvz)}[\nabla_{\alpha_i}f_{\alpha_i}(\rvz_i)] \nonumber
\end{eqnarray}
By applying Eqn.\ref{normalization-grad-3} and Eqn.\ref{normalization-grad-beta}, we have
\begin{eqnarray}
 \label{z-generator-grad-appendix}
\nabla_{\beta_i}\log p_{\theta}(\rvx)  &=&\E_{p_{\theta}(\rvz|\rvx)}[\nabla_{\beta_i}\log p_{\beta_i}(\rvz_i|\rvz_{i+1})] \\
&-& \E_{p_{\alpha, \beta_{>0}}(\rvz)}[\nabla_{\beta_i}\log p_{\beta_i}(\rvz_i|\rvz_{i+1})] \nonumber
\end{eqnarray}
\subsection{Variational Learning}\label{sec:appendix-variational}
Recall that $L(\theta, \omega) = D_{\mathrm{KL}}(q_{\omega}(\rvx,\rvz)||p_{\theta}(\rvx,\rvz))$. We can view such joint KL as a surrogate of the MLE objective with the KL perturbation term, i.e., $L(\theta, \omega) = D_{\mathrm{KL}}(p_{\rm data}(\rvx)||p_\theta(\rvx)) + D_{\mathrm{KL}}(q_\omega(\rvz|\rvx)||p_\theta(\rvz|\rvx))$. Specifically, we have

\begin{eqnarray*}
&&D_{\KL}(p_{\rm data}(\rvx)||p_{\theta}(\rvx)) + D_{\KL}(q_{\omega}(\rvz|\rvx)||p_{\theta}(\rvz|\rvx))\\
&=&-\mathbb E_{p_{\rm data}}[\log p_\theta(\rvx)] + D_{\mathrm{KL}}(q_\omega(\rvz|\rvx) | p_{\theta}(\rvz|\rvx))+C\\
&=& \mathbb E_{p_{\rm data}} \left[ \mathbb E_{q_\omega(\rvz|\rvx)}\left( \log \frac{q_\omega(\rvz|\rvx)}{p_\theta(\rvz|\rvx)}\right) 
-  \log p_\theta(\rvx) \right] +C \\
&=& \E_{p_{\rm data}}  \left[- \E_{q_{\omega}(\rvz|\rvx)}\left[\frac{p_{\theta}(\rvx, \rvz)}{q_{\omega}(\rvz|\rvx)}\right]\right] + C \\
&=& \E_{p_{\rm data}} [-\tilde{L}(\theta,\omega)] + C
\end{eqnarray*}
where $C \equiv - H(p_{\rm data}(x))$ is the entropy of the empirical data distribution and can be treated as constant. $\tilde{L}(\theta,\omega)$ is a lower bound of the log-likelihood $\log p_{\theta}(\rvx)$ typically known as ELBO \cite{kingma2013auto}. Notice that, with the joint EBM prior model, we consider the KL optimization between the aggregate posterior and EBM prior model, i.e., $\tilde{L}(\theta,\omega) = \E_{q_{\omega}(\rvz|\rvx)} [\log p_{\beta_0}(\rvx|\rvz)] - D_{\KL}(q_{\omega}(\rvz|\rvx)||p_{\alpha, \beta_{>0}}(\rvz))$, while VAEs compute $D_{\KL}(q_{\omega}(\rvz|\rvx)||p_{\beta_{>0}}(\rvz))$, where $p_{\beta_{>0}}(\rvz)$ is the Gaussian prior model.

Therefore, we can compute the gradient $\nabla_{\theta,\omega}\tilde{L}(\theta,\omega)$ to jointly update the inference, generator and EBM prior model. Learning the prior model $(\alpha_i, \beta_{>0})$ involves computing the derivative of $\log \mathrm{Z}_{\alpha, \beta_{>0}}$, which can be referred to Eqn.\ref{normalization-grad-2} and Eqn.\ref{normalization-grad-3}.

\subsection{Change of Variable}\label{sec:appendix-cov}
We observe that using Langevin dynamic on latent space for deep hierarchical structures can be heterogeneous, where latent variables may be formed in different shapes (e.g., spatial variables and vectors) and can rely on the distribution that has a high variance. Therefore, we further consider $\epsilon_{\rvz}$-space, which has a unit variance and can make the prior sampling more efficient and effective. For brevity, we take a two-layer structure as an example, i.e., $\rvz = (\rvz_1, \rvz_2)$, where for L layers, the derivation is the same. 

\noindent \textbf{Deterministic transformation $T_{\beta_{>0}}$:} For generator model $p_{\beta_{>0}}(\rvz_1, \rvz_2)$, $\rvz_1$ follows conditional Gaussian distribution as $p(\rvz_1|\rvz_2) \sim \N (\mu_{\beta_1}(\rvz_2), \sigma_{\beta_1}(\rvz_2))$, while $p(\rvz_2)$ is assumed to be unit Gaussian, such that $p(\rvz_2) \sim \N(0, I_d)$. Let $(\epsilon_{\rvz_1}, \epsilon_{\rvz_2})$ be the re-parametrization variables, we have $T_{\beta_{>0}}$ defined as
\begin{eqnarray}
\label{t_1}
\rvz_2 &=& T_{\beta_{>0}}^{\rvz_2}(\epsilon_{\rvz_2}) = \epsilon_{\rvz_2}\\
\label{t_2}
\rvz_1 &=& T_{\beta_{>0}}^{\rvz_1}(\epsilon_{\rvz_1}, \epsilon_{\rvz_2}) = \mu_{\beta_1}(\rvz_2) + \sigma_{\beta_1}(\rvz_{2}) \cdot\epsilon_{\rvz_1}
\end{eqnarray}
$T_{\beta_{>0}}^{\rvz_2}(\epsilon_{\rvz_2})$ and $T_{\beta_{>0}}^{\rvz_1}(\epsilon_{\rvz_1}, \epsilon_{\rvz_2})$ are invertible and usually referred as reparameterization trick used in VAEs. Thus, the re-parametrization variables $(\epsilon_{\rvz_1}, \epsilon_{\rvz_2})$ can be independently drawn from Gaussian noise, i.e., $(\epsilon_{\rvz_{1}}, \epsilon_{\rvz_{2}}) \sim p_{\epsilon}(\epsilon_{\rvz_{1}}, \epsilon_{\rvz_{2}})$, where $p_{\epsilon}(\epsilon_{\rvz_{1}}, \epsilon_{\rvz_{2}}) = p_{\epsilon_1}(\epsilon_{\rvz_{1}})p_{\epsilon_2}(\epsilon_{\rvz_{2}})$ and $p_{\epsilon_i}(\epsilon_{\rvz_{i}}) \sim \N(0, I_{||\epsilon_{\rvz_i}||})$.

\noindent \textbf{Toward $\epsilon_{\rvz}$-space $p_{\alpha, \beta_{>0}}(\epsilon_{\rvz_1}, \epsilon_{\rvz_2})$:} With invertible transformation $T_{\beta_{>0}}$, we can apply change of variable rule as
\begin{eqnarray}
\label{t_3}
p_{\beta_{>0}}(\rvz_1, \rvz_2) &=& p_{\epsilon}(\epsilon_{\rvz_{1}}, \epsilon_{\rvz_{2}}) | \mathrm{det} (J_{T_{\beta_{>0}}^{-1}})| \\
\label{t_4}
p_{\epsilon}(\epsilon_{\rvz_{1}}, \epsilon_{\rvz_{2}}) &=& p_{\beta_{>0}}(\rvz_1, \rvz_2) | \mathrm{det} (J_{T_{\beta_{>0}}})|
\end{eqnarray}
where $J_{T_{\beta_{>0}}}$ is the Jacobian of $T_{\beta_{>0}}$. 

For brevity, we denote $\epsilon_\rvz =(\epsilon_{\rvz_{1}}, \epsilon_{\rvz_{2}})$, then $p_{\beta_{>0}}(\rvz) = p_{\epsilon}(\epsilon_{\rvz}) | \mathrm{det} (J_{T_{\beta_{>0}}^{-1}})|$ and $p_{\epsilon}(\epsilon_{\rvz}) = p_{\beta_{>0}}(\rvz) | \mathrm{det} (J_{T_{\beta_{>0}}})|$. Recall that the proposed joint EBM prior model is defined as $p_{\alpha,\beta_{>0}}(\rvz)$.  With change of variable, $p_{\alpha, \beta_{>0}}(\epsilon_{\rvz})$ is
\begin{equation}
\begin{aligned}
\label{COV}
&p_{\alpha, \beta_{>0}}(\epsilon_{\rvz}) = p_{\alpha, \beta_{>0}}(\rvz) | \mathrm{det} (J_{T_{\beta_{>0}}})| \nonumber\\
&= \frac{1}{\mathrm{Z}_{\alpha, \beta_{>0}}}\exp f_{\alpha}(T_{\beta_{>0}}(\epsilon_{\rvz}))p_{\beta_{>0}}(\rvz)| \mathrm{det} (J_{T_{\beta_{>0}}})|\nonumber\\
&= \frac{1}{\mathrm{Z}_{\alpha, \beta_{>0}}}\exp f_{\alpha}(T_{\beta_{>0}}(\epsilon_{\rvz}))p_{\epsilon}(\epsilon_{\rvz}) \nonumber
\end{aligned}
\end{equation}
Therefore, sampling from $p_{\alpha,\beta_{>0}}(\rvz)$ can be done by first sampling $\epsilon_{\rvz}$ from $p_{\alpha,\beta_{>0}}(\epsilon_{\rvz})$ and then using deterministic transformation $T_{\beta_{>0}}$ to obtain $\rvz$ as Eqn.\ref{t_1} and Eqn.\ref{t_2}. Compared to latent space $p_{\alpha,\beta_{>0}}(\rvz)$, the $\epsilon_{\rvz}$-space $p_{\alpha,\beta_{>0}}(\epsilon_{\rvz})$ independently draws samples from the same Gaussian distribution, and such distribution has a unit variance allowing us to use the fixed step size of Langevin dynamic to efficiently and effectively explore the latent space at different layers for deep hierarchical structures. For experiments with backbone model BIVA \cite{maaloe2019biva} or NVAE \cite{vahdat2020nvae}, we adopt similar reparametrized sampling scheme as VAEBM \cite{xiao2020vaebm} via public code\footnote{https://github.com/NVlabs/VAEBM}.
\section{Additional Experiments}

\subsection{Hierarchical Representations}\label{sec:appendix-hierarchical}
\begin{figure}[h]
\centering
\includegraphics[width=0.92\columnwidth]{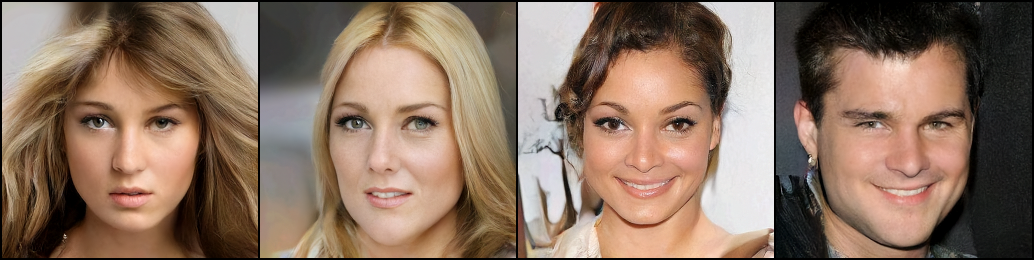}
\includegraphics[width=0.92\columnwidth]{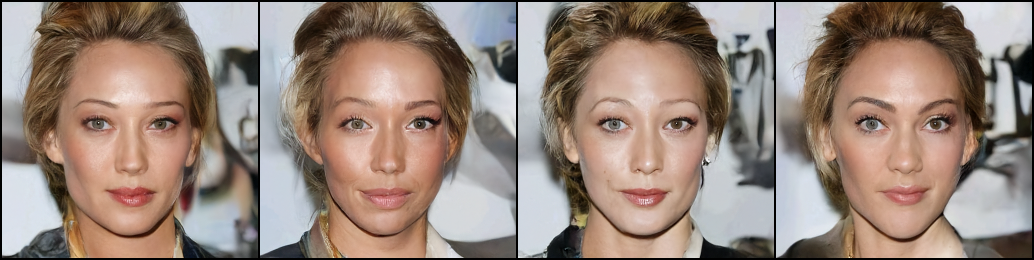}
\includegraphics[width=0.92\columnwidth]{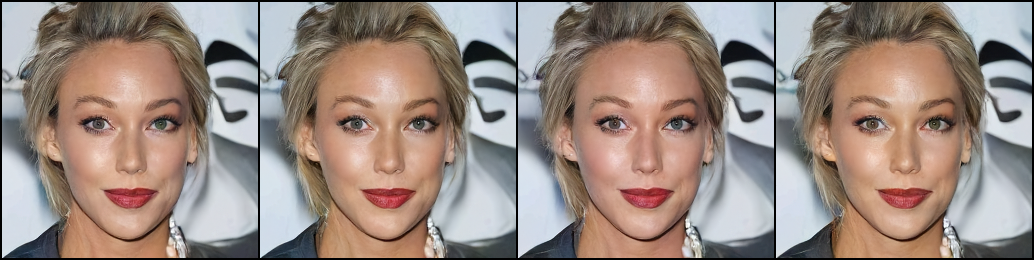}
\and\caption{Hierarchical sampling with NVAE backbone on CelebA-HQ-256.}
\label{Fig.nvae_hs}
\end{figure}

\noindent\textbf{Hierarchical reconstruction.} To examine the hierarchical representation, we further conduct hierarchical reconstruction by replacing the inferred latent vectors at the bottom layers with the ones from the prior distribution. We use BIVA \cite{maaloe2019biva} as our backbone model for multi-layer generator and inference model, and we use Langevin dynamic for prior sampling. Specifically, we run prior Langevin sampling for the latent codes at lower layers (e.g., $\rvz_{i \le k}$)  with the latent codes at top layers (from BIVA inference model) remaining fixed (using Eqn.20 in main text). We train our model on CelebA-64 and show hierarchical reconstructions in Fig.\ref{Fig.hierarchical-rec}. 

We observe that the details in reconstructions can be gradually replaced by common features as more layers of latent variables are sampled from the prior distribution. For example, the sunglass first becomes a more common glass and then eventually disappears. This concurs with the observation in \cite{havtorn2021hierarchical}, suggesting that our model carries different levels of abstract representations within the hierarchical structure.

\begin{figure}[h]
    \centering
    \begin{subfigure}{.2\columnwidth}
    \centering
        \includegraphics[height=1cm,]{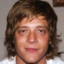}
        \includegraphics[height=1cm,]{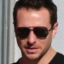}
    \subcaption{Example.}
    \end{subfigure}\hspace{-7mm}
    \begin{subfigure}{.78\columnwidth}
    \centering
    \includegraphics[height=1cm,]{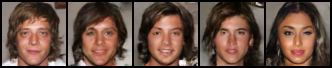}
    \includegraphics[height=1cm,]{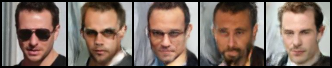}
    \subcaption{Sampling from bottom layer to top layer.}
    \end{subfigure}
    \caption{Hierarchical reconstruction}
    \label{Fig.hierarchical-rec}
\end{figure}

\noindent \textbf{Additional results for OOD detection:} In addition, we compute AUROC, AUPRC and FPR80 for BIVA and our EBM prior model in OOD detection. We use the log-likelihood $L^{>k}$ and a ratio type $LLR^{>k}$ \cite{havtorn2021hierarchical} as the decision functions for BIVA. If the low-level representations are well-learned at the bottom layers, using decision function with higher k should render better detection performance for reducing impact of shared low-level features. The results are shown in Tab.\ref{table.ood}.
\begin{table}[h]
\centering
\resizebox{0.99\columnwidth}{!}{
\begin{tabular}{c | c c c} 
\toprule
BIVA / Ours & AUROC$\uparrow$  & AUPRC$\uparrow$ & FPR80$\downarrow$ \\ 
\midrule
 $\text{L}^{>0}$ / $\text{L}^{>0}_{\text{EBM}}$ & 0.066 / 0.087 & 0.339 / 0.319 & 0.997 / 0.999\\
 $\text{L}^{>3}$ / $\text{L}^{>3}_{\text{EBM}}$ & 0.307 / 0.324 & 0.427 / 0.438 & 0.970 / 0.972\\
 $\text{L}^{>6}$ / $\text{L}^{>6}_{\text{EBM}}$ & 0.436 / 0.449 & 0.514 / 0.528 & 0.942 / 0.942 \\
 $\text{L}^{>9}$ / $\text{L}^{>9}_{\text{EBM}}$ & 0.866 / 0.870 & 0.855 / 0.858 & 0.230 / 0.227\\
 $\text{LLR}^{>9}$ / $\text{LLR}^{>9}_{\text{EBM}}$& 0.885 / \textbf{0.927} & 0.876 / \textbf{0.918} & 0.200 / \textbf{0.113}\\
\bottomrule
\end{tabular}%
}
\caption{AUROC, AUPRC and FPR80 for BIVA and our EBM prior model on CIFAR10(in) / SVHN(out).}
\label{table.ood}
\end{table}


\section{Experiment Details}

\noindent \textbf{Fr$\acute{e}$chet Inception Distance:} We compute FID scores with 30,000 generated images for CelebA-HQ-256 and 50,000 generated images for other data.

\noindent \textbf{Implementations:} For comparisons in generator models with informative prior, we train our model on SVHN (32 x 32), CIFAR-10 (32 x 32), and CelebA-64 (64 x 64), where we use full training split of SVHN and CIFAR-10 and 40,000 cropped training examples of CelebA-64 following the protocol in \cite{pang2020learning}. All training images are resized and scaled to [-1, 1]. For applying to NVAE backbone models, we train our joint EBM prior on latent variables of all layers. The implementations of models on CelebA-64 and EBMs for NVAE backbone are shown in Tab.\ref{table.celeba64_joint_structure}. We denote the operation of convolution and transposed convolution as $\rm conv (k, c, s)$ and $\rm convT (k, c, s)$, where $\rm k$ is the kernel size, $\rm c$ is the channel number and $\rm s$ is the stride number, and we denote $\rm LeakyReLU$ as $\rm LReLU$.

\begin{table}[t]
\centering
\begin{tabular}{|c || c |} 
\hline
Layers & In-Out Size \\
\midrule
\multicolumn{2}{|c|}{EBM $f_{\alpha_i}(\rvz_i)$ for NVAE backbone} \\
\midrule
Input: $\rvz_i$ & (h x w x c) \\
N x conv (4, 64, 2), LReLU & (4 x 4 x 64) \\
N x Linear (200), LReLU & 200 \\
Linear (1) & 1 \\
\bottomrule
\midrule
\multicolumn{2}{|c|}{Generator Model $p_{\beta_1}(\rvz_1|\rvz_2)$} \\
\midrule
Input: $\rvz_2$ & 100 \\
Linear (200), LReLU & 200 \\
Linear (200), LReLU & 200 \\
Linear (200) & 200 \\
Split for $\mu_{\rvz_1}$and $\log \sigma_{\rvz_1}$ & 100, 100\\
\midrule
\multicolumn{2}{|c|}{Generator Model $p_{\beta_0}(\rvx|\rvz)$} \\
\midrule
Input: $\rvz_1$ & (1 x 1 x 100)\\
convT (4, 1024, 1), LReLU & (4 x 4 x 1024)\\
convT (4, 512, 2), LReLU & (8 x 8 x 512) \\
convT (4, 256, 2), LReLU & (16 x 16 x 256)\\
convT (4, 128, 2), LReLU & (32 x 32 x 128)\\
convT (4, 3, 2), Tanh & (64 x 64 x 3)\\
\midrule
\multicolumn{2}{|c|}{Inference Model $q_{\omega_2}(\rvz_2|\rvz_1)$} \\
\midrule
Input: $\rvz_1$ & 100 \\
Linear (200), LReLU & 200 \\
Linear (200), LReLU & 200 \\
Linear (200) & 200 \\
Split for $\mu_{\rvz_2}$and $\log \sigma_{\rvz_2}$ & 100, 100\\
\midrule
\multicolumn{2}{|c|}{Inference Model $q_{\omega_1}(\rvz_1|\rvx)$} \\
\midrule
Input: $\rvx$ & (64 x 64 x 3)\\
conv (4, 128, 2), LReLU & (32 x 32 x 128)\\
conv (4, 256, 2), LReLU & (16 x 16 x 256) \\
conv (4, 512, 2), LReLU & (8 x 8 x 512)\\
conv (4, 1024, 2), LReLU & (4 x 4 x 1024)\\
conv (4, 200, 1) & (1 x 1 x 200)\\
Split for $\mu_{\rvz_1}$and $\log \sigma_{\rvz_1}$ & 100, 100\\
\midrule
\multicolumn{2}{|c|}{EBM $f_{\alpha_1}(\rvz_1)$} \\
\midrule
Input: $\rvz_1$ & 100 \\
Linear (200), LReLU & 200 \\
Linear (200), LReLU & 200 \\
Linear (200), LReLU & 200 \\
Linear (200), LReLU & 200 \\
Linear (1) & 1 \\
\midrule
\multicolumn{2}{|c|}{EBM $f_{\alpha_2}(\rvz_2)$} \\
\midrule
Input: $\rvz_2$ & 100 \\
Linear (100), LReLU & 100 \\
Linear (100), LReLU & 100 \\
Linear (1) & 1 \\
\bottomrule
\end{tabular}
\caption{Network structures for generation, inference and EBMs on CELEBA-64 and EBM structure for NVAE backbone models. }
\label{table.celeba64_joint_structure}
\end{table}

\section{Additional qualitative results:}\label{sec:appendix-additional} We show additional image synthesis for CIFAR-10, LSUN-Church-64 and CelebA-HQ-256 in Fig.\ref{Fig.syn_cifar10}, Fig.\ref{Fig.syn_church}, Fig.\ref{Fig.syn_256_0.7} and Fig.\ref{Fig.syn_256_1.0}. The additional visualizations of langevin transition that starts from $p_{\beta_{>0}}(\rvz)$ toward the learned EBM prior distribution $p_{\alpha,\beta_{>0}}(\rvz)$ are shown in Fig.\ref{Fig.ld_cifar10}, Fig.\ref{Fig.ld_church} and Fig.\ref{Fig.ld_256}.

\begin{figure*}[b]
\centering
\includegraphics[width=0.98\textwidth]{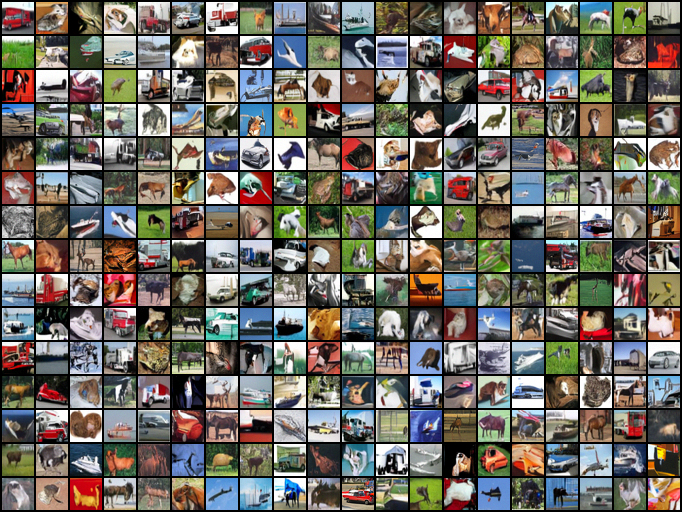}
\and\caption{Generated images on CIFAR-10. Samples are uncurated.}
\label{Fig.syn_cifar10}
\end{figure*}

\begin{figure*}[b]
\centering
\includegraphics[width=0.49\textwidth]{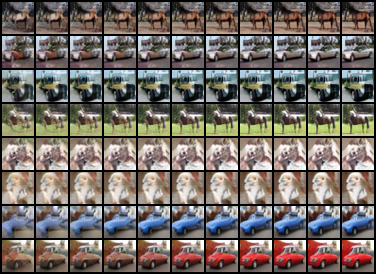}
\includegraphics[width=0.49\textwidth]{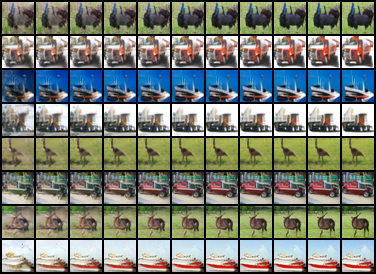}
\and\caption{Langevin transition on CIFAR-10.}
\label{Fig.ld_cifar10}
\end{figure*}

\begin{figure*}[b]
\centering
\includegraphics[width=0.8\textwidth]{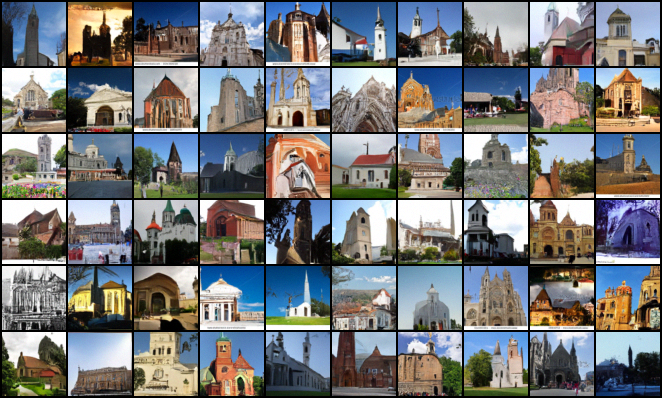}
\and\caption{Generated images on LSUN-Church-64. Samples are uncurated.}
\label{Fig.syn_church}
\end{figure*}

\begin{figure*}[b]
\centering
\includegraphics[width=0.8\textwidth]{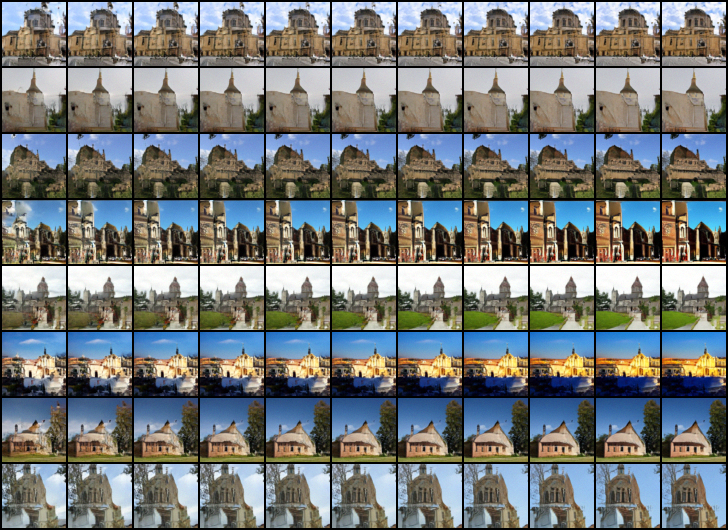}
\and\caption{Langevin transition on LSUN-Church-64.}
\label{Fig.ld_church}
\end{figure*}

\begin{figure*}[b]
\centering
\includegraphics[width=0.85\textwidth]{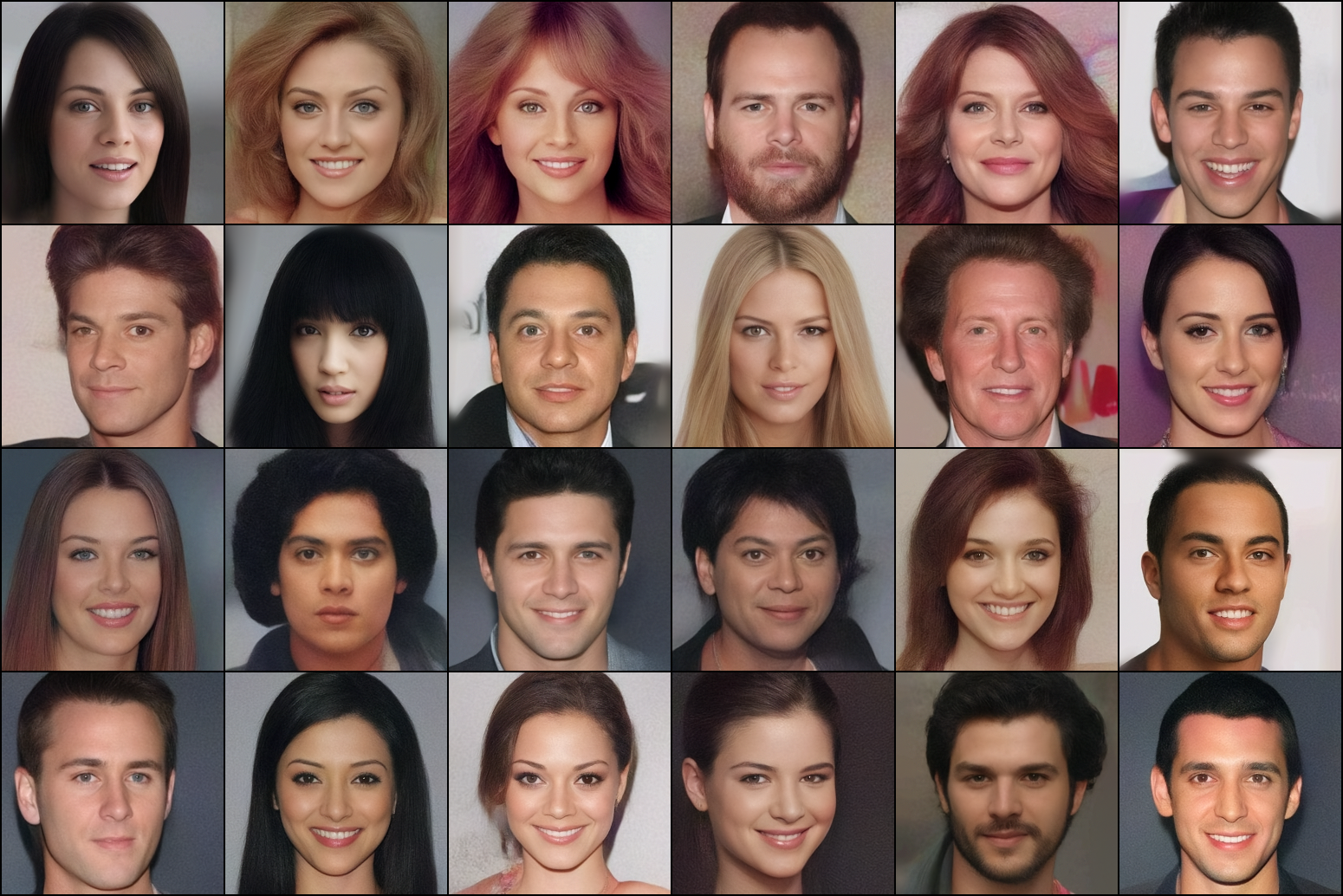}
\and\caption{Generated images on CelebA-HQ-256 (temperature t=0.7). Samples are uncurated.}
\label{Fig.syn_256_0.7}
\end{figure*}

\begin{figure*}[b]
\centering
\includegraphics[width=0.85\textwidth]{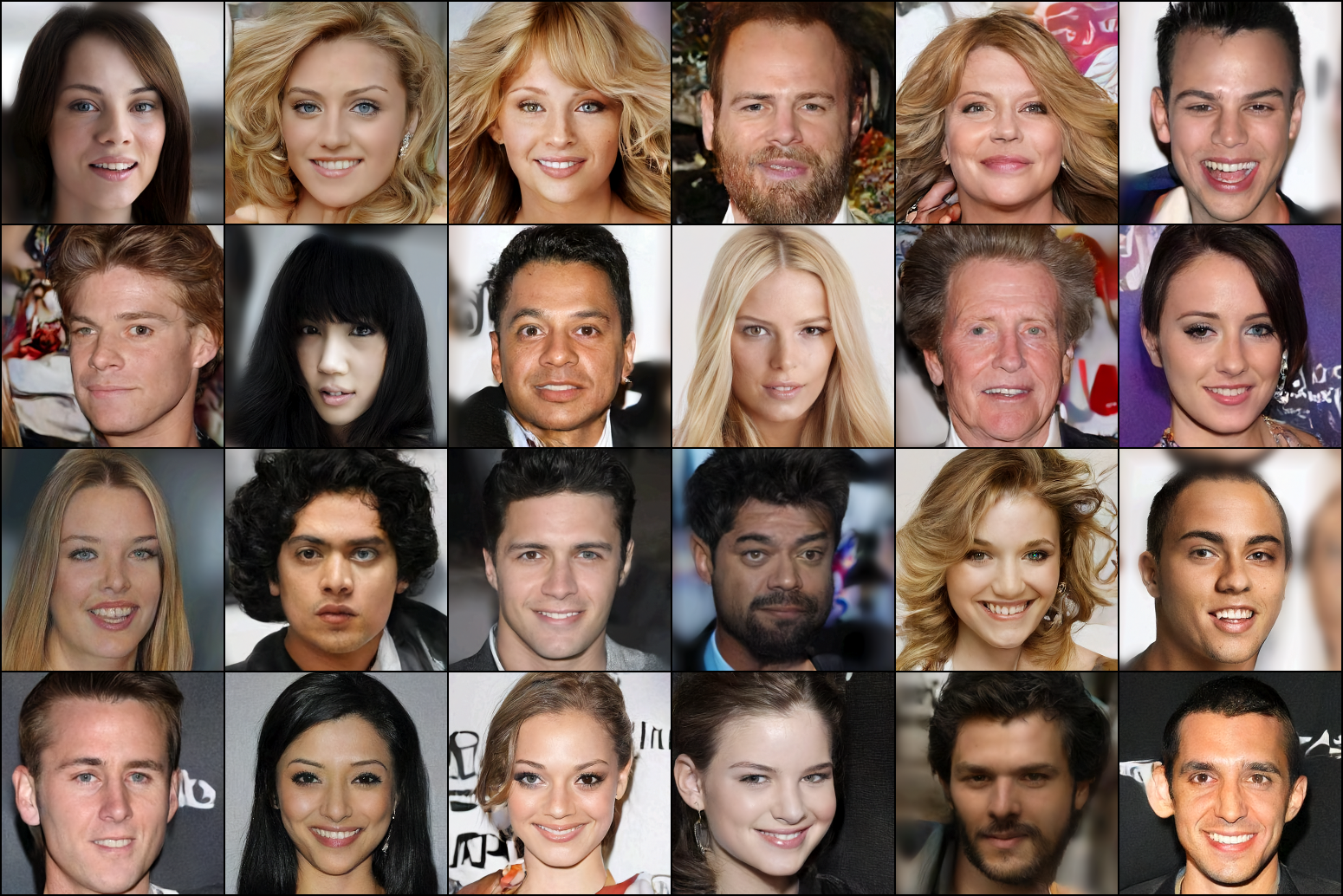}
\and\caption{Generated images on CelebA-HQ-256 (temperature t=1.0). Samples are uncurated.}
\label{Fig.syn_256_1.0}
\end{figure*}

\begin{figure*}[b]
\centering
\includegraphics[width=0.8\textwidth]{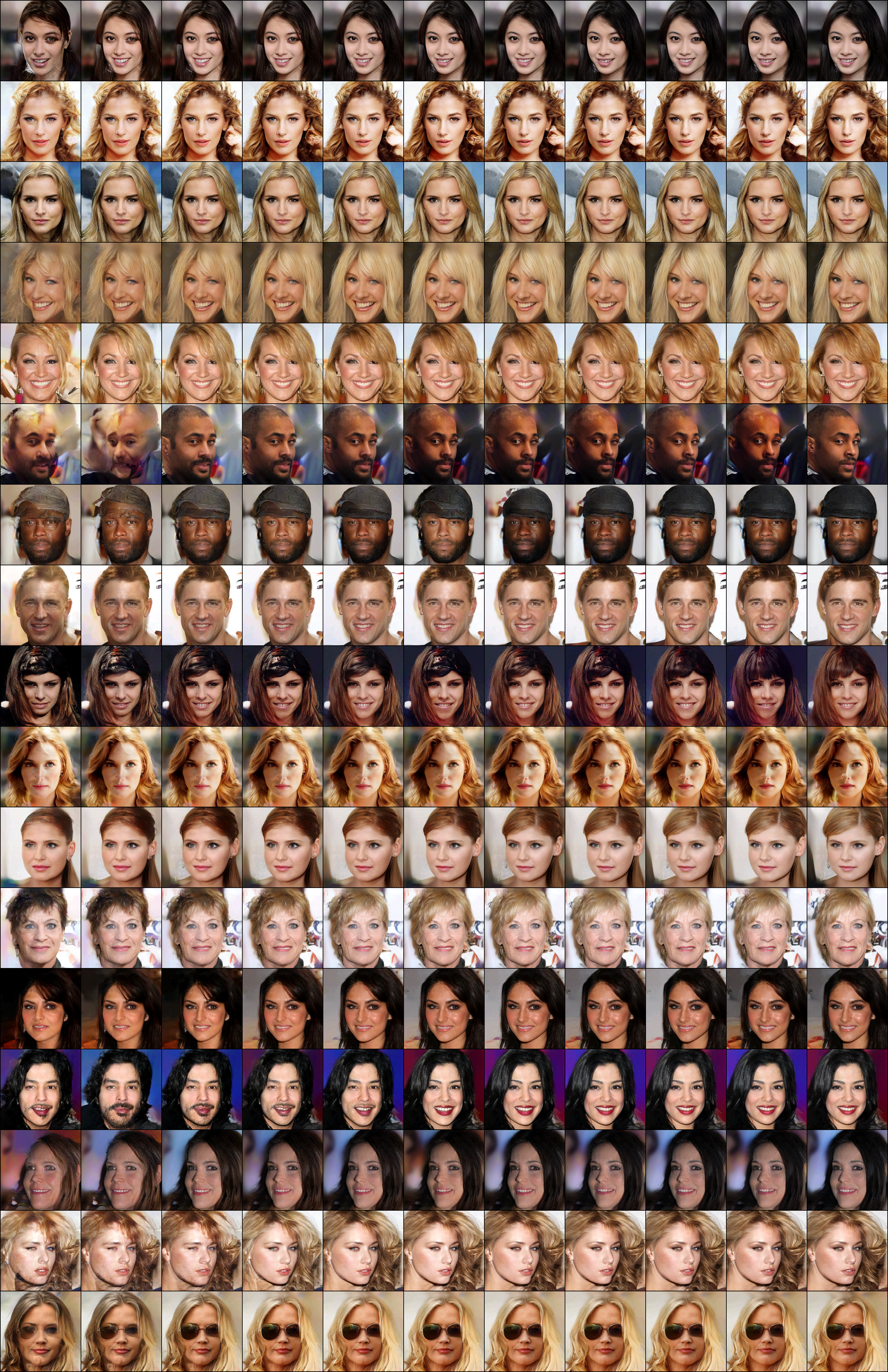}
\and\caption{Langevin transition on CelebA-HQ-256. (temperature t=1.0).}
\label{Fig.ld_256}
\end{figure*}

\end{alphasection}
\end{document}